\documentclass[runningheads]{llncs}
\usepackage{graphicx}
\usepackage{amice_shortcuts}
\usepackage{lipsum}
\usepackage{tikz}
\usetikzlibrary{patterns}
\usepackage{bibunits}
\usepackage{subcaption}
\usepackage{hyperref}
\usepackage{leftidx}
\hypersetup{
    colorlinks=true,
    linkcolor=black,
    citecolor = black,
    filecolor=magenta,      
    urlcolor=blue
    }
\usepackage{float}
\usepackage[linesnumbered, ruled, lined, boxed, commentsnumbered]{algorithm2e}

\tikzstyle{EDB}=[draw=white, opacity = 0.0,line width=2pt, preaction={clip, postaction={pattern=north west lines, pattern color=black}}]

\makeatletter
\newcommand{\printfnsymbol}[1]{%
  \textsuperscript{\@fnsymbol{#1}}%
}
\newcommand\blfootnote[1]{%
  \begingroup
  \renewcommand\thefootnote{}\footnote{#1}%
  \addtocounter{footnote}{-1}%
  \endgroup
}
\makeatother
\defaultbibliographystyle{IEEEtran} 
\defaultbibliography{IEEEabrv,bibliographies/amice_refs,bibliographies/refs}

\title{Finding and Optimizing Certified, Collision-Free Regions in Configuration Space for Robot Manipulators}

\author{Alexandre Amice\thanks{equal contribution}\inst{1} \and
Hongkai Dai\printfnsymbol{1}\inst{2}\and
Peter Werner\inst{1}\and
Annan Zhang\inst{1}\and
Russ Tedrake\inst{1,2}}

\institute{MIT \email{\{amice, wernerpe, annanz, russt\}@mit.edu}\and Toyota Research Institute \email{\{hongkai.dai\}@tri.global}}

\date{February 2022}

\authorrunning{Amice, Dai, et al.}
\titlerunning{Certified, Collision-Free Regions in C-Space}

\begin{document}

\maketitle

\begin{abstract}
   Configuration space (C-space) has played a central role in collision-free motion planning, particularly for robot manipulators. While it is possible to check for collisions at a point using standard algorithms, to date no practical method exists for computing collision-free C-space \emph{regions} with rigorous certificates due to the complexities of mapping task-space obstacles through the kinematics. In this work, we present the first to our knowledge method for generating such regions and certificates through convex optimization. Our method, called C-{\sc{Iris}} (C-space Iterative Regional Inflation by Semidefinite programming),  generates large, convex polytopes in a rational parametrization of the configuration space which are guaranteed to be collision-free. Such regions have been shown to be useful for both optimization-based and randomized motion planning. Our regions are generated by alternating between two convex optimization problems: (1) a simultaneous search for a maximal-volume ellipse inscribed in a given polytope and a certificate that the polytope is collision-free and (2) a maximal expansion of the polytope away from the ellipse which does not violate the certificate. The volume of the ellipse and size of the polytope are allowed to grow over several iterations while being collision-free by construction. Our method works in arbitrary dimensions, only makes assumptions about the convexity of the obstacles in the \emph{task} space, and scales to realistic problems in manipulation. We demonstrate our algorithm's ability to fill a non-trivial amount of collision-free C-space in a 3-DOF example where the C-space can be visualized, as well as the scalability of our algorithm on a 7-DOF KUKA iiwa and a 12-DOF bimanual manipulator.
   \blfootnote{This work was supported by Office of Naval Research, Award No. N00014-18-1-2210, National Science Foundation, Award No. EFMA-1830901., and Air Force Research Lab Award No. FA8750-19-2-1000}
\end{abstract}

\begin{bibunit}
\section{Introduction and Related Work}

\begin{figure}[htb]
\centering
\begin{subfigure}{0.45\textwidth}
\includegraphics[width=0.9\textwidth]{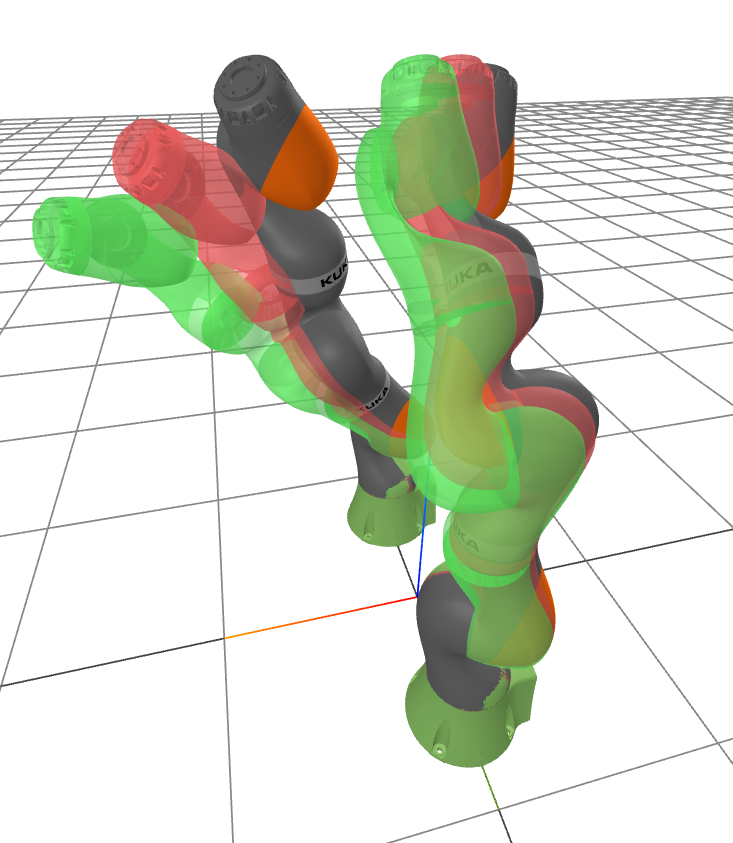}
\end{subfigure}
\begin{subfigure}{0.45\textwidth}
\includegraphics[width=0.9\textwidth]{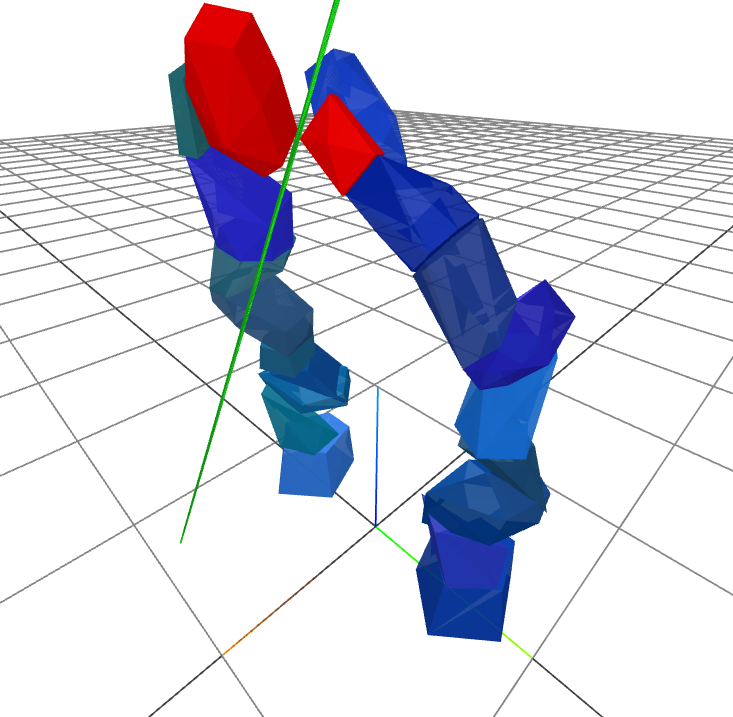}
\end{subfigure}
\caption{\small{Left: Multiple configurations sampled from one certified C-free region on a 12-DOF dual KUKA iiwa systems (with wrist joints fixed). Each configuration is visualized using one color. Right: Our certified C-free region is tight. At one of the sampled configurations, we draw the separating plane  (green) as the non-collision certificate between the two highlighted polytopic collision geometries (red), with a distance of 7.3mm. Every link of the iiwa is approximated by one or several polytopes.}}
\label{Fig: dual_iiwa}
\vspace{-13.5pt}
\end{figure}

The notion of configuration space (C-space) has become a foundational idea in robot motion planning since its proposal in the seminal work \cite{lozano1983spatial}. In the presence of obstacles in the Cartesian task space, a fundamental challenge is describing the collision-free C-space (C-free): the full range of configurations for which a robot is not in collision. Prior work has sought to describe a C-space obstacle from its task space description \cite{kavraki1995computation}\cite{branicky1990computing}. However it is recognized that a complete, mathematical description of C-space obstacles with high degree of freedom systems is intractable \cite{canny1988complexity}. 
% perhaps drop this sentence so we can refer to Fig 1 before Fig 2
% \red{This complexity can be seen in Fig \ref{F: collision constraint pinball} where we visualize the configuration space obstacle of a simple 3-DOF robot.}

\begin{figure*}
    \centering
    \begin{subfigure}[c]{0.19\textwidth}
    \centering
    \makebox[0pt]{
        \includegraphics[width = 2.3\textwidth, trim={15cm 6cm 17cm 6cm},clip]{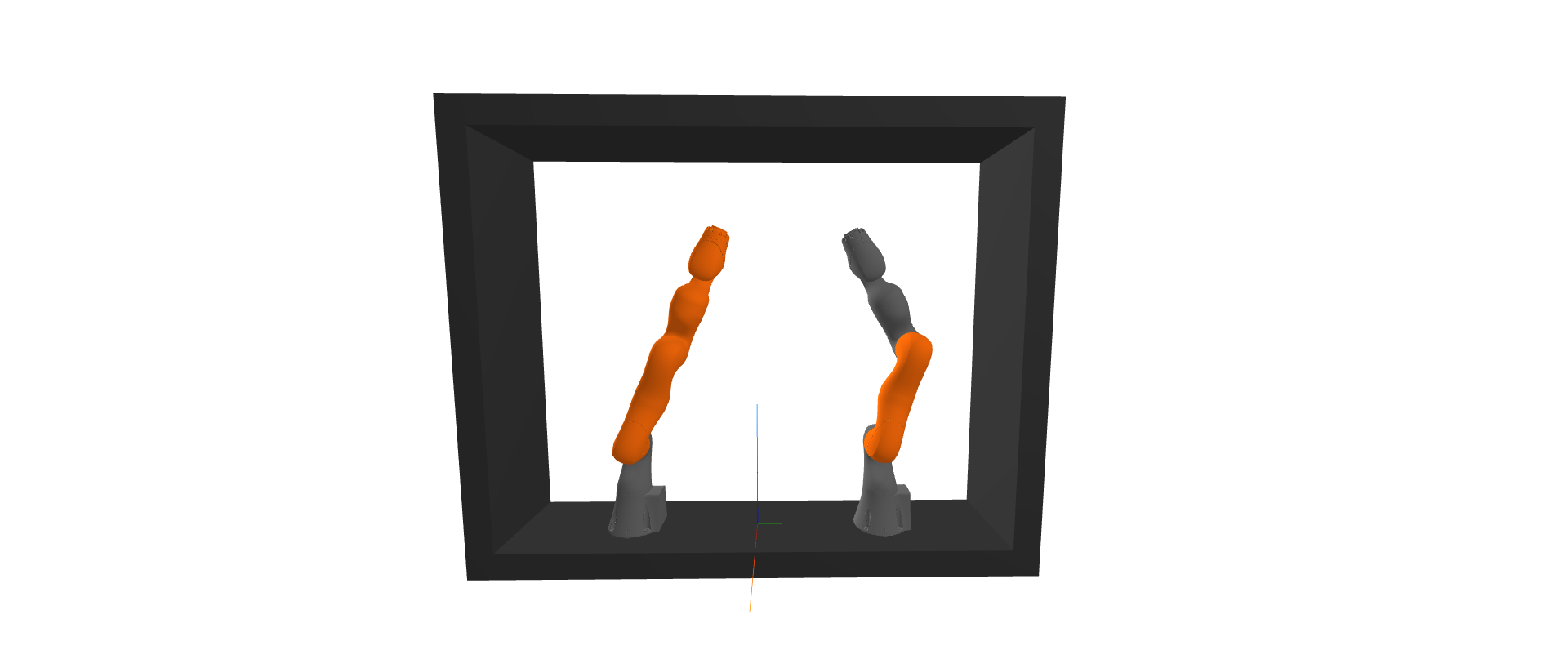}
        %\caption{Start of Trajectory} \label{F: cert cfg2}
        }
    \end{subfigure}
    \hspace{1em}
    \begin{subfigure}[c]{0.33\textwidth}
    \centering
        \includegraphics[width = 1.4\textwidth, trim={20cm 0cm 6cm 1cm},clip]{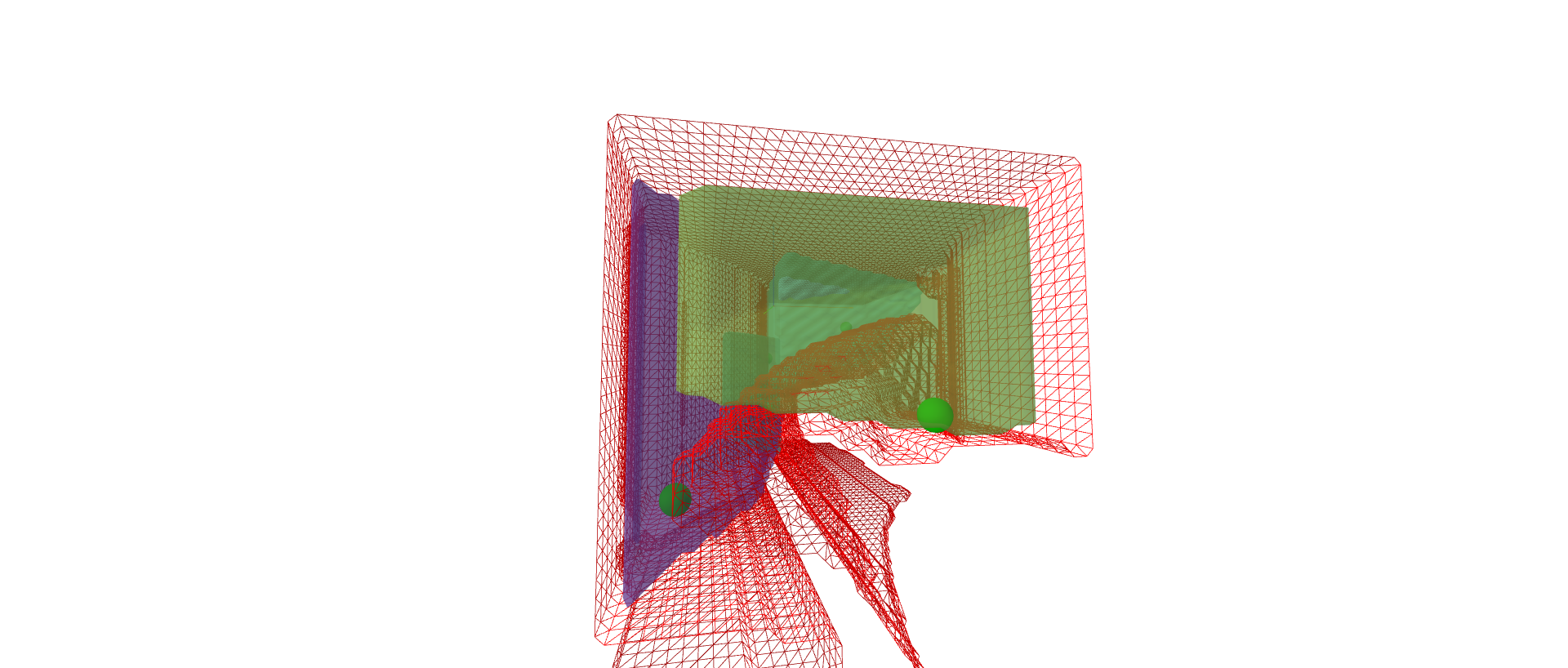}
        %\caption{Start of Trajectory} \label{F: cert cfg2}
    \end{subfigure}
    \begin{subfigure}[c]{0.34\textwidth}
    \centering
        \includegraphics[width = 1.3\textwidth, trim={11cm 1cm 11cm 3cm},clip]{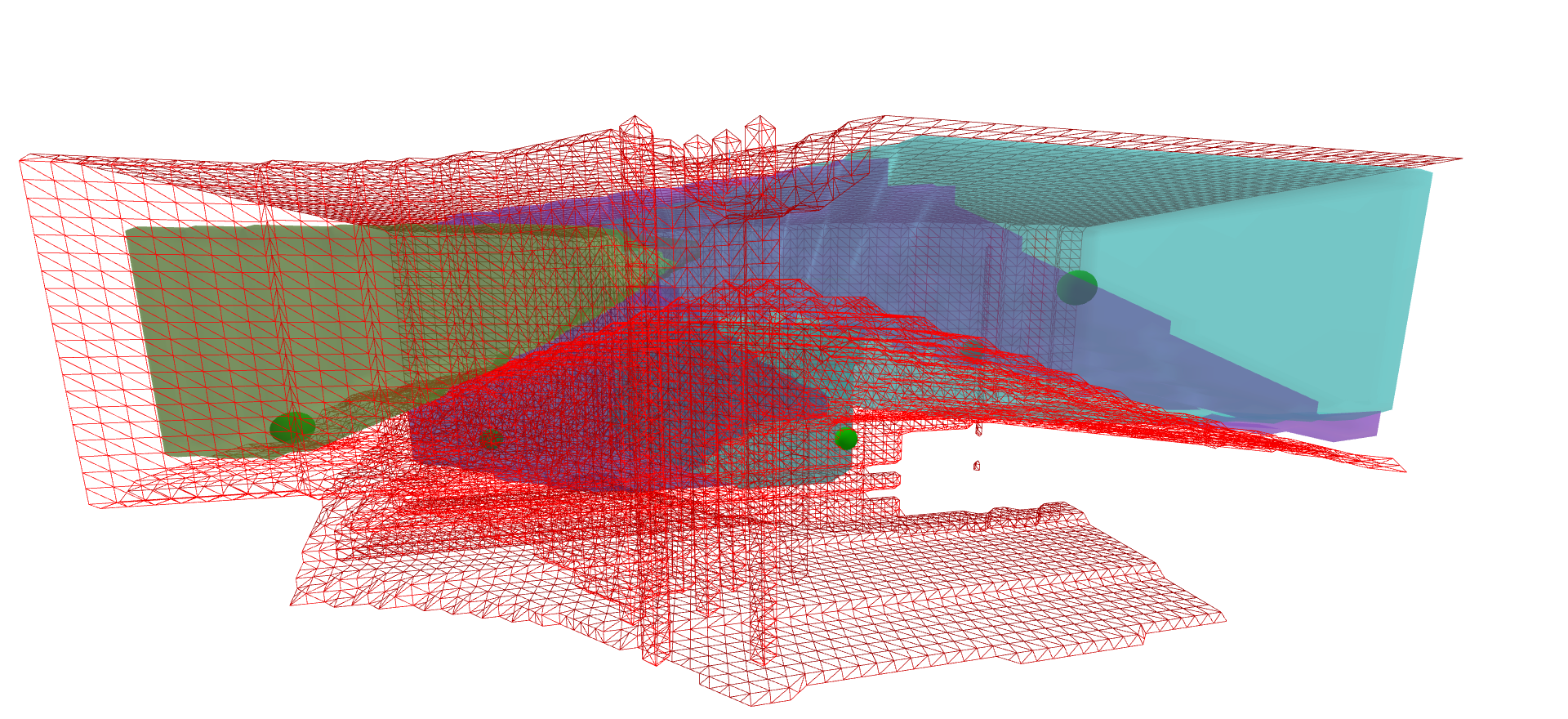}
        %\caption{Middle of Trajectory} \label{F: cert cfg1}
    \end{subfigure}
    \caption{The 3-DOF robotic system described in Section \ref{S: Pinball}. In the right hand panels we visualize the C-space of this robot. The bright red mesh denotes the collision constraint of the robot with itself and with the outer box. The free space is given by the interior of the mesh and is filled by the polytopic regions certified by Algorithm \ref{A: Bilinear Alternation} and grown from the seed points in bright green.}
    \label{F: collision constraint pinball}
    \vspace{-13.5pt}
    %%%%%%
    % \centering
    % \begin{subfigure}[t]{0.33\textwidth}
    % \centering
    %     \includegraphics[width = 1.0\textwidth, trim={15cm 4cm 15cm 4cm},clip]{figures/pinball_iiwas_no_grid.png}
    %     %\caption{Start of Trajectory} \label{F: cert cfg2}
    % \end{subfigure}
    % \begin{subfigure}[t]{0.33\textwidth}
    % \centering
    %     \includegraphics[width = 1.0\textwidth, trim={20cm 0cm 6cm 1cm},clip]{figures/cspace_filled_region1.png}
    %     %\caption{Start of Trajectory} \label{F: cert cfg2}
    % \end{subfigure}
    % \begin{subfigure}[t]{0.32\textwidth}
    % \centering
    %     \includegraphics[width = 1.0\textwidth, trim={11cm 1cm 11cm 3cm},clip]{figures/cspace_filled_region2.png}
    %     %\caption{Middle of Trajectory} \label{F: cert cfg1}
    %     \vspace{ 0.3 cm}
    % \end{subfigure}
    % \caption{The 3-DOF robotic system described in Section \ref{S: Pinball}. In the right hand panels we visualize the C-space of this robot. The bright red mesh denotes the collision constraint of the robot with itself and with the outer box. The free space is given by the interior of the mesh and is filled by the polytopic regions certified by Algorithm \ref{A: Bilinear Alternation} and grown from the seed points in bright green.}
    % \label{F: collision constraint pinball}
    % \vspace{-13.5pt}
\end{figure*}

Despite this difficulty, describing C-free has been the subject of many works in robotics. Randomized, collision-free motion planners such as Rapidly Exploring Random Trees (RRT) \cite{lavalle1998rapidly}, Probabilistic Road Maps (PRM) \cite{kavraki1996probabilistic}, and their variants can be seen as attempting to describe C-free via piecewise linear paths defined by the nodes and edges of the graph. Works such as \cite{verghese2022configuration}\cite{han2019configuration}\cite{wong2014adaptive} also seek to provide a description of C-free via randomized sampling. All of these methods guarantee that the configurations at the sample points are collision-free and attempt to provide a probabilistic certificate that nearby points contain no collisions. Unfortunately, such methods are inherently limited by the density of the samples used, which can become a barrier in higher dimensions. This work will overcome this barrier by providing a deterministic certificate that all of the infinitely many configurations in a given subset of C-space are collision-free. This certificate comes in the form of one parametric separating hyperplane for each pair of objects which can collide in a given environment such as the one seen in Fig \ref{Fig: dual_iiwa}. C-free will then be described as a union of these certified sets. Fig \ref{F: collision constraint pinball} shows an example of the complexity C-space obstacles for even a simple 3-DOF robot (the red mesh), as well as an example of several certified, collision-free polytopic regions produced by our method.

Describing C-free as a union of sets is not a new idea. In two and three dimensions and in the presence of polyhedral obstacles, this problem is equivalent to describing a non-convex polyhedron as a union of convex sets. It is known that finding a minimal such decomposition is NP-hard \cite{lingas1982power} to solve exactly and even APX-hard \cite{eidenbenz2003approximation} to approximate\footnote{A problem is said to be APX-hard if no polynomial time algorithm can achieve an approximation ratio of $1+\delta$ for some $\delta > 0$ unless $P = NP$.}. Works such as \cite{lien2007approximate} and \cite{ghosh2013fast} overcome these hardness results by finding decompositions that are unions of approximately convex sets. 

A method of describing C-free in arbitrary dimensions is given by the {\sc{Iris}} algorithm in \cite{deits2015computing}. Under the assumption of known, convex obstacles in C-space, {\sc{Iris}} can rapidly grow convex, polytopic regions by alternating between two convex programs. The collision-free certificates of {\sc{Iris}} arise naturally due to the assumption of convexity of the obstacles. Unfortunately, it is often the case that obstacles are naturally described as convex sets in \emph{task space} which are rarely convex in C-space.

Searching for certificates of non-collision in C-space when obstacles are specified as convex sets in the task space using convex programming (specifically Sum-Of-Squares (SOS) programming) is the primary technical contribution of this work. This is achieved by a novel formulation exploiting the well-known algebraic structure of the kinematics \cite{diankov2010automated}\cite{trutman2020globally}\cite{sommese2005numerical}. Similar to \cite{deits2015computing}, we construct certified, collision-free polytopic regions by alternating between a pair of convex programs. Our method works in arbitrary dimensions and is the first to our knowledge to provide such certificates in this setting.
 
Describing C-free as a union of convex regions is particularly attractive as such descriptions have proven useful for optimization-based motion planning such as in \cite{deits2015efficient}\cite{schouwenaars2001mixed}. Indeed, in our companion paper \cite{marcucci2022motion}, the authors demonstrate great success in finding the globally-optimal collision-free trajectory for robot manipulators by planning through the types of regions generated by our method.

Our certified, convex regions also find natural use in randomized motion planners. By paying an upfront cost to describe C-free as a union of polytopic sets, piecewise-linear randomized motion planners can certify that both individual samples and edges connecting those samples are completely collision-free by simply checking membership in a finite number of polytopes.

We begin Section \ref{S: Problem Formulation} by describing how non-collision of a single configuration can be certified via convex programming. In Section \ref{S: Background}, we give essential background needed to describe the technical approach described in Section \ref{S: Technical Approach}, where we extend the key ideas of the problem formulation to handle a range of configurations. Section \ref{S: Technical Approach} culminates in Algorithm \ref{A: Bilinear Alternation} where we give our algorithm C-IRIS for growing certified regions. We begin in Section \ref{S: Results} by exploring a simple 3-DOF robot in which we can visualize the configuration space and follow by demonstrating the ability of our algorithm to certify a wide range of postures for a realistic 7-DOF robot. We conclude by showing our algorithm's ability to scale by exploring a 12-DOF, bimanual manipulator.

\noindent\textbf{Notation:} 
Throughout the paper, we will use calligraphic letters ($\calS$) to denote sets, Roman capitals ($X$) to denote matrices and Roman lower case ($x$) to denote vectors. We use $[N] = \{1, \dots, N\}$ and we will denote the cone of Sums-of-Squares (SOS) polynomials as $\bSigma$. Additionally, we will adopt the notation of \cite{tedrakeManip} for rigid transforms.% Finally, given a multivariate polynomial $f(x)$ we will distinguish between its total degree $d$, which is the maximum sum of the exponents appearing in a single term, and its coordinate degree which is the maximum exponent of each variable which appears in each term. For example $f(x,y) = x^3y^2+xy^6$ has total degree $7$ but coordinate degree $3$ and $6$ for variables $x$ and $y$ respectively.

\section{Problem Formulation} \label{S: Problem Formulation}
We assume a known environment in which both the robot and obstacles in the \emph{task space} have been decomposed into a union of compact, convex, vertex-representation (V-rep) polytopes. Such polytopic collision geometries of task space are readily available through standard tools (e.g. V-HACD).

Our objective is to find large, convex regions of C-free regardless of the dimension of the configuration space. This objective is beyond the scope of current decomposition methods such as V-HACD due to the complexity of the non-linear kinematics of the robot and the dimensionality of interesting problems

\begin{figure}[htb]
    \centering
    \scalebox{0.90}{
    \input{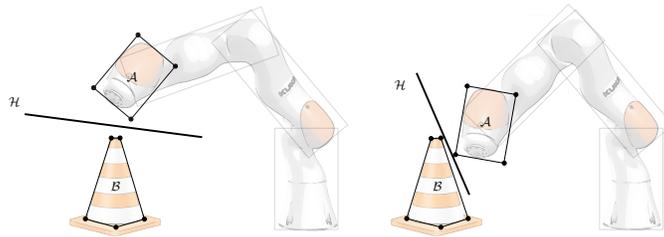}
    }
    \caption{The convex collision geometries $\mathcal{A}$ and $\mathcal{B}$ are collision-free if and only if there exists a separating plane $\mathcal{H}$. The plane acts as a certificate of non-collision.}
    \label{fig:svm}
    \vspace{-13.5pt}
\end{figure}

We begin by considering the problem of certifying that two collision geometries $\calA$ and $\calB$ in task space are non-intersecting for a fixed configuration. Recall that two convex bodies are non-intersecting if and only if there exists a plane $\calH$ separating the two bodies \cite{boyd2004convex}. Moreover, any point in $\calA$ and $\calB$ can be written as a convex combination of the vertices of $\calA$ and $\calB$, respectively. Therefore, a finite certificate that $\calA$ and $\calB$ are not in collision is to find a plane $a^Tx + b = 0$ such that all the vertices $\calA_{j}$ lie on one side of the plane while the vertices $\calB_{k}$ lie on the other side as in Fig \ref{fig:svm}. Adopting the convention established in \cite{tedrakeManip}, we denote the position of vertex $\calA_{j}$ in a given frame $F$ as $\leftidx{^F}p^{\calA_{j}}$. The non-collision condition is certified by the existence of a separating plane satisfying the following linear constraints on parameters $a, b$ \cite{bishop2006pattern}:
\begin{align}
\begin{aligned}
    &~ a^T\left(\leftidx{^F}p^{\calA_{j}}\right) + b \geq 1 ~\forall~ j 
    \\
    &~ a^T\left(\leftidx{^F}p^{\calB_{k}}\right) + b \leq -1 ~\forall~ k.
    \end{aligned}
    \label{E: svm}
\end{align}
This hyperplane proves that all the points in body $\calA$ are separated from $\calB$ by a distance of at least $\frac{2}{\norm{a}}$. We emphasize that for each pair of collision geometries we need to search for a separating plane for that pair.

While such a formulation provides a certificate for a single configuration, it is not sufficient for an entire set of possible configurations, as different configurations might require different separating planes as seen in Fig \ref{fig:svm}. Therefore, we will search for a family of separating planes where $a(q), b(q)$ are parameterized functions of the robot configuration $q$. We search for these parameters through optimization such that the separating plane condition \eqref{E: svm} holds for any configuration within a certain C-space region $\calP$, namely
\begin{align}
     \forall q\in\calP \implies \begin{cases}
    a(q)^T\left(\leftidx{^F}p^{\calA_{j}}(q)\right) + b(q) \geq 1 ~\forall~ j 
    \\
     a(q)^T\left(\leftidx{^F}p^{\calB_{k}}(q)\right) + b(q) \leq -1 ~\forall~ k
    \end{cases},
    \label{E: imply_separating}
\end{align}
where $\implies$ means the condition on the left implies the condition on the right.

The right hand side of \eqref{E: imply_separating} are non-negativity conditions $\left(a^T\right)\leftidx{^F}p^{\calA_j}+b-1\ge 0$ and $ -1-\left(a^T\right)\leftidx{^F}p^{\calB_k}-b \ge 0$. It is generally challenging if not intractable to verify a non-negativity condition for the infinitely many element in a set ($\forall q\in\calP$ on the left hand side of \eqref{E: imply_separating}). But when the non-negativity condition and the set have certain structures, then the verification can become tractable.
% One such structure are polynomial functions.
As we will see in Section \ref{S: Background}, when both the set $\calP$ and the non-negativity conditions on both side of $\implies$ are specified by polynomials, we can solve a convex optimization program (specifically a Sum-of-Squares program) to verify the condition. In Section \ref{S: Technical Approach} we will convert \eqref{E: imply_separating} to polynomial functions so as to verify the non-collision condition with the Sum-of-Squares technique.

%\begin{remark}
% We remark that the condition in \eqref{E: svm} is necessary and sufficient to certify non-collision between \emph{any} two convex geometries (e.g., spheres, cylinders, etc) based on their extreme points. We choose to simplify the presentation by focusing on V-rep polytopes as they provide a convenient method of providing a finite set of necessary and sufficient constraints for such a separating plane. This assumption is in fact not essential.
%\end{remark}

\section{Background} \label{S: Background}
As hinted in Section \ref{S: Problem Formulation} and as we shall show in Section \ref{S: Technical Approach}, the problem of certifying a C-free region can be posed as a polynomial implication problem of the form
\begin{align} \label{E: Gen Cert Prob}
    \forall x \in \calS_{g} = \{x: g_{i}(x) \geq 0, i \in [n]\} \implies p(x) \geq 0,
\end{align}
where $g_i(x), p(x)$ are all polynomials of $x$. While checking the positivity of a polynomial is in general NP-hard \cite{parrilo2000structured}, a sufficient condition is to check whether $p(x)$ can be expressed as a sum of squares $p(x) = \sum_{i} f_{i}^{2}(x)$, with $f_i(x)$ also being polynomials. Such a check can be performed using semidefinite programming (SDP) \cite{parrilo2000structured}\cite{blekherman2012semidefinite} and is known as Sums-of-Squares (SOS) programming. The SOS technique has been widely used in robotics, for example in stability verification \cite{tedrake2010lqr}\cite{majumdar2017funnel}\cite{shen2020sampling}, reachability analysis \cite{jarvis2003some}\cite{yin2021backward} and geometric modelling \cite{ahmadi2016geometry}.

Moreover, polynomial implications of the type in \eqref{E: Gen Cert Prob} are well studied in algebraic geometry \cite{blekherman2012semidefinite} and indeed algebraic certificates of the implication can be searched for using SOS programming \cite{parrilo2000structured}. Theorems concerning these implications are typically called Positivstellensatz (Psatz) theorems \cite{stengle1974nullstellensatz}. One of the strongest theorems concerning implications of the form in \eqref{E: Gen Cert Prob} is known as Putinar's Positivstellensatz:

\begin{theorem}[Putinar's Positivstellensatz \cite{putinar1993positive}] \label{T: Putinar}
Suppose $\calS_{g}$ is Archimedean. Then $p(x) > 0$ for all $x \in \calS_{g}$ if and only if there exists $\lambda_{i}(x)$ SOS such that:
\begin{align} 
    p(x) = \lambda_{0}(x) + \sum_{i} \lambda_{i}(x)g_{i}(x) \tag{C2} \label{E: C2}.
\end{align}
The polynomials $\lambda_{i}(x)$ for $i \geq 0$ are referred to as multiplier polynomials.
\end{theorem} 
where Archimedean is a property slightly stronger than compactness.  It is defined in Appendix \ref{A: Archimedean}.

In order to apply the SOS technique and Theorem \ref{T: Putinar} to our non-collision condition \eqref{E: imply_separating}, we need to convert the point positions $\leftidx{^F}p^{\calA_j}, \leftidx{^F}p^{\calB_k}$ to polynomial functions. While the point positions are typically written as trigonometric functions of $q$ using $\sin$ and $\cos$, we will build upon a strong history in robotics of algebraic kinematics, e.g. \cite{sommese2005numerical}\cite{wampler1990numerical}, to write the kinematics function using polynomials in Section \ref{S: Rat Forward}.

\section{Technical Approach} \label{S: Technical Approach}
Given an initial, collision-free posture specified as joint angles in the configuration space, our algorithm searches for two key objects: a C-free region $\calP$ and a certificate $\calC_{\calP}$ that $\calP$ contains only collision-free postures. In Section \ref{S: Rat Forward} we will introduce a rational parameterization of the robot configuration. This parameterization enables us to write the non-collision condition \eqref{E: imply_separating} as a polynomial implication problem, certifiable through solving a Sum-of-Squares program (Section \ref{S: Certification}). In Section \ref{S: Bilinear Alternation}, we build on the program from Section \ref{S: Certification} to enable the search for regions increasing in size. This is outlined in Algorithm \ref{A: Bilinear Alternation}.

\subsection{Rational Parametrization of the Forward Kinematics} \label{S: Rat Forward}
In order to write the non-collision condition \eqref{E: imply_separating} as a polynomial implication problem \eqref{E: Gen Cert Prob}, we need to convert both sides of $\implies$ in \eqref{E: imply_separating} to polynomials. To this end, we will utilize a rational reparameterization of robot configurations, such that the position of a point on the robot, computed through forward kinematics, is converted to a rational function (with both the numerator and the denominator as polynomials of this reparameterization). This enables us to use Theorem \ref{T: Putinar} to verify the non-collision condition in Section \ref{S: Certification} using SOS.

We consider a robot manipulator with $N$ revolute joints\footnote{Our approach can be extened to robots with algebraic joints, including revolute, prismatic, cylindrical, plane and spherical joints \cite{wampler2011numerical}.}. The position of a point $A$ expressed in the reference frame $F$ as a function of the joint angles $q$ generically assumes the form of the trigonometric polynomial
\begin{align}
    \leftidx{^F}p^{A}_w(q) = 
     \sum_{j}
    c_{jw}\prod_{i \in \calS_{F,A}} \left(\cos^{n_{ij}}(q_{i})\sin^{m_{ij}}(q_{i}) \right),
    \; w\in\{x, y, z\},
    \label{E: gen trig poly}
\end{align}
where $\calS_{F,A} \subseteq [N]$ is the set of joints lying on the kinematic chain between $F$ and $A$, $m_{ij} + n_{ij} \leq 1$, and both $m_{ij}$ and $ n_{ij}\in\{0, 1\}$ for all $i, j$ \cite{craig2005introduction}. Namely, for each $q_i$, at most one of $\cos(q_i)$ or $\sin(q_i)$ can appear in $\prod(\cos^{n_{ij}}(q_i)\sin^{m_{ij}}(q_i)$; hence $\cos(q_1)\sin(q_2)\cos(q_3)$ and $\sin(q_1)\cos(q_3)$ are allowed, but $\cos(q_1)\sin(q_1)$ or $\cos^2(q_1)$ are not. $c_{jx}, c_{jy}, c_{jz}$ are given constants computed from the D-H parameters of the kinematics chain. We choose to be explicit about the reference frame $F$ at the risk of being pedantic, as the choice of reference frame $F$ will have important consequences for the scalability of the approach described in Section \ref{S: Certification} (see Appendix \ref{S: Frame Selection} for a detailed discussion).

The function in \eqref{E: gen trig poly} is known as a multi-linear trigonometric polynomial. It has many fortunate algebraic properties which we will exploit in subsequent sections. One particular property that we shall use is a change of variables known as the stereographic projection \cite{spivakCalc}:
\begin{align} \label{E: rational sub}
    s_{i} &= \tan\left( \frac{q_{i}}{2}\right)\rightarrow
    \begin{cases}
    \cos(q_{i}) =\frac{1-s_{i}^{2}}{1+s_{i}^{2}} \\
    \sin(q_{i}) = \frac{2s_{i}}{1+s_{i}^{2}}
    \end{cases} .
\end{align}
We will refer to the space defined by the variable $q$ as the configuration space and the space defined by the variable $s$ as the tangent-configuration space.

In this new variable $s$, the forward kinematics are given by the following rational function, where both the numerator and denominator are polynomial functions of $s$,
\begin{align}
    \leftidx{^F}p^{A}_w(s) = \sum_jc_{jw}\prod_{i \in \calS_{F,A}} \left(\frac{(1-s_i^2)^{n_{ij}}(2s_i)^{m_{ij}}}{(1+s_i^2)^{n_{ij}+m_{ij}}}\right), w\in\{x, y, z\}.
    \label{E: gen rat poly}
\end{align}
This parametrization and the theorems in Section \ref{S: Background} will be leveraged in Section \ref{S: Certification} to transform the certification problem \eqref{E: imply_separating} into a SOS program.

Our objective will be to find large, polytopic regions $\calP = \{s \mid Cs \leq d\}$ in the variables $s = \tan(q/2)$. We will assume that the joints of the robot cannot undergo complete rotation and are constrained to angles
\begin{align}
    -\pi < q_{l} \leq q \leq q_{u} < \pi \label{E: joint limits}.
\end{align}
This ensures that the mapping between the configuration and tangent-configuration spaces is bijective and so trajectories in the tangent-configuration space correspond unambiguously to trajectories in the original configuration space. Moreover, it allows us to restrict the polytope $\calP \subseteq \calP_{\lim} = \{s \mid s_{l} \leq s \leq s_{u}\} = \{s \mid C_{\lim} s \leq d_{\lim}\}$ the polytope encoding the joint limits. This ensures that $\calP$ is a compact polytope and hence we can invoke both directions of Theorem \ref{T: Putinar} due to \cite[Corrollary 6.3.5]{prestel2013positive}.
\subsection{The SOS Certification Problem} \label{S: Certification}
We certify the non-collision condition \eqref{E: imply_separating} through solving a SOS program which we describe in this section. The key idea is to generalize \eqref{E: svm} and search for a polynomial family of separating planes parametrized by the tangent-configuration space variable $s$ which separate two collision geometries $\calA$ and $\calB$ for all $s$ in a given polyhedron $\calP$. These separating planes serve as a certificate that $\calA$ and $\calB$ are not in collision for any configurations in $\calP$. By simultaneously searching for a plane for each collision pair $(\calA, \calB)$ in the environment, we are able to certify that no collision occurs between any two bodies for all $s \in \calP$.

To begin, we denote the task-space-polytope vertex position from \eqref{E: gen rat poly} as a vector of rational functions
\begin{align*}
    \leftidx{^F}p^{\calA_{j}}(s) = \frac{\leftidx{^F}f^{\calA_{j}}(s)}{\leftidx{^F}g^{\calA_{j}}(s)}.
\end{align*}

Notice that $\leftidx{^F}g^{\calA}(s) = \prod_{i\in\mathcal{S}_{F,A}} (1+s_{i}^{2})^{n_{ij}+m_{ij}}$ is a product of positive polynomials and therefore positive for all $s$. Moreover, this denominator is common for all x/y/z entries of $\leftidx{^F}p^{\calA}(s)$. By plugging $\leftidx{^F}p^{\calA}(s)$ into the constraints of \eqref{E: svm} and multiplying the denominator through, we are able to write a pair of \emph{polynomial} inequalities analogous to the constraints in \eqref{E: svm} 
\begin{subequations}
\begin{align}
    &\qquad\underset{\forall~\text{pairs} (\calA, \calB)}{\text{find }}
    a_{\calA, \calB}(s),~b_{\calA, \calB}(s) ~\subjectto
    \\
    &\underbrace{~a_{\calA, \calB}^{T}(s)\left(\leftidx{^F}f^{\calA_{j}}(s)\right) + (b_{\calA, \calB}(s)-1)\left(\leftidx{^F}g^{\calA_{j}}(s)\right)}_{\alpha^{F, \calA_{j}}(a_{\calA, \calB}, b_{\calA, \calB}, s)}  \geq 0 ~\forall~ j, ~s \in \calP \label{E: alpha poly}
    \\
    &\underbrace{-a_{\calA, \calB}^{T}(s)\left({}^{F}f^{\calB_{k}}(s)\right) - (b_{\calA, \calB}(s)+1)\left({}^{F}g^{\calB_{k}}(s)\right)}_{\beta^{F, \calB_{k}}(a_{\calA, \calB}, b_{\calA, \calB}, s)}\geq 0  ~\forall~ k , ~s \in \calP\label{E: beta poly},
    \end{align}
    \label{E: single pair cert}
\end{subequations} 
where we denote the polynomials on the left hand side of \eqref{E: alpha poly}\eqref{E: beta poly} as $\alpha^{F, \calA_{j}}(a_{\calA, \calB}, b_{\calA, \calB}, s)$ and $\beta^{F, \calB_{k}}(a_{\calA, \calB}, b_{\calA, \calB}, s)$ for convenience. $a_{\calA,\calB}(s), b_{\calA,\calB}(s)$ are also polynomials of $s$. The $\alpha^{F, \calA_{j}}(a_{\calA, \calB}, b_{\calA, \calB}, s)$ and $\beta^{F, \calB_{k}}(a_{\calA, \calB}, b_{\calA, \calB}, s)$ are linear in the coefficients of the polynomials $a_{\calA, \calB}$ and  $b_{\calA, \calB}$ which are the decision variables in the problem. The variables $s$ are known as the indeterminates of the polynomial.

At this point, we recognize that the non-collision condition \eqref{E: imply_separating} can be written as the following polynomial implication:
\begin{align*}
    % \forall 
    s \in \calP=\{s \mid Cs \leq d\} \implies \eqref{E: alpha poly} \text{ and } \eqref{E: beta poly}.
\end{align*}

Using Theorem \ref{T: Putinar} and \cite[Corrollary 6.3.5]{prestel2013positive}, a necessary and sufficient condition for this implication to hold is the existence of polynomials $\lambda_{i}^{\calA_{j}, \calB}(s), \nu^{\calA, \calB_{k}}_{i}(s) \in \bSigma$ (where $\bSigma$ is the set of SOS polynomials) such that
\begin{subequations}
\begin{align}
    \alpha^{F, \calA_{j}}(a_{\calA, \calB}, b_{\calA, \calB}, s)= 
    \lambda_{0}^{\calA_{j},\calB}(s) + \sum_{i=1}^{m}\lambda_{i}^{\calA_{j}, \calB}(s)(d_{i} -c_{i}^Ts)
    \label{E: Putinar pos side} 
    \\
    \beta^{F, \calB_{k}}(a_{\calA, \calB}, b_{\calA, \calB}, s) =
    \nu^{\calA, \calB_{k}}_{0}(s) + \sum_{i=1}^{m}\nu^{\calA, \calB_{k}}_{i}(s)(d_{i} -c_{i}^Ts)
    \label{E: Putinar neg side},
\end{align}
\end{subequations}
where $c_i, d_i$ are the i-th row of $C, d$. The equality between the polynomials in \eqref{E: Putinar pos side}\eqref{E: Putinar neg side} means that the coefficients of the corresponding terms are equal, which adds linear equality constraints on the decision variables $a_{\calA,\calB}, b_{\calA, \calB}, \lambda^{\calA_j,\calB}, \mu^{\calA,\calB_k}$.

Therefore, a polytopic region in the tangent-configuration space $\calP = \{s \mid Cs \leq d\}$ is collision-free if and only if the program
\begin{gather}
    \begin{gathered}
    \underset{\forall~\text{pairs} (\calA, \calB)}{\text{find }}
    a_{\calA, \calB}(s),~b_{\calA, \calB}(s), ~ \lambda(s), \nu(s) ~\subjectto
    \\
    \alpha^{F, \calA_{j}}(a_{\calA, \calB}, b_{\calA, \calB}, s) = 
    \lambda_{0}^{\calA_{j},\calB}(s) + \sum_{i=1}^{m}\lambda_{i}^{\calA_{j}, \calB}(s)(d_{i} -c_{i}^Ts)
    \\
    \beta^{F, \calB_{k}}(a_{\calA, \calB}, b_{\calA, \calB}, s) =
    \nu^{\calA, \calB_{k}}_{0}(s) + \sum_{i=1}^{m}\nu^{\calA, \calB_{k}}_{i}(s)(d_{i} -c_{i}^Ts)
    \\
    \lambda(s), \nu(s) \in \bSigma
    \end{gathered}
    \label{E: cert problem}
\end{gather}
is feasible, where $\lambda(s)$ and $\nu(s)$ collect all the multipliers $\lambda_{i}^{\calA_{j}, \calB}$ and $\nu_{i}^{\calA, \calB_{k}}$. This is a SOS program which searches for the coefficients of the polynomials $a_{\calA, \calB}(s),~ b_{\calA, \calB}(s),~ \lambda(s)$, and $\nu(s)$. We call a feasible solution 
\begin{align*}
    \calC_{\calP} = \bigcup_{\calA, \calB}\{a_{\calA, \calB}(s),~ b_{\calA, \calB}(s),~ \lambda(s), \nu(s)\}
\end{align*} 
to \eqref{E: cert problem} a collision-free certificate of $\calP$.

\begin{remark}
If $\calP$ is collision-free, then it can always be certified as such by \eqref{E: cert problem} provided the degree of $\alpha$ and $\beta$ are sufficiently large. We discuss practical considerations of basis selection for \eqref{E: cert problem} in Appendix \ref{S: Basis Selection}

\end{remark}

\subsection{Growing Polytopic Regions in Tangent-Configuration Space} \label{S: Bilinear Alternation}
Describing C-free using a few large regions rather than many smaller ones is advantageous in almost all circumstances. In this section, we show how to extend the SOS feasibility program in \eqref{E: cert problem} to enable not only certification, but certified \emph{growth} of polytopic C-free regions.

% To facilitate using our algorithm to efficiently cover the configuration space with certified regions, for instance as part of an approximate convex decomposition, we would like to certify a few large regions rather than many small ones. In this section we show how to extend the SOS feasibility program in \eqref{E: cert problem} into one which can also grow the certified region in the C-space.

We begin by discussing how we will measure the size of our polytope $\calP=\{s \mid Cs\le d\}$. While it may be attractive to measure the size of a polytope by its volume, it is known that computing the volume of a half-space representation (H-Rep) polytope is \#P-hard\footnote{\#P-hard problems are at least as hard as NP-complete problems \cite{provan1983complexity}} \cite{dyer1988complexity} and therefore intractable as an objective. A useful surrogate for the volume of $\calP$ used in \cite{deits2015computing} is the volume of the maximal inscribed ellipse of $\calP$, the set $\calE_{\calP} = \{Qu + s_{0}\mid \norm{u}_2 \leq 1\}$ where $Q$ describes the shape of the ellipsoid and $s_{0}$ its center. The problem of finding the maximal inscribed ellipsoid in a given polytope is a semidefinite program described in \cite[Section 8.4.2]{boyd2004convex}.

Additionally, in order to cover diverse areas of C-free, we grow each polytope $\calP$ around some nominal posture $s_{s}$ we call the seed point. New seed points are typically chosen using rejection sampling to obtain a point outside of the existing certified regions and $\calP$ is required to contain $s_{s}$ as it grows.

% To encourage coverage of C-free, we seed each new region $\calP$ with a sample point, $s_{s}$ (using rejection sampling to obtain a point outside of the existing certified regions), and add constraint \eqref{E: cert with ellipse contain sample} that $s_{s}$ must be contained in $\calP$. 

% \red{This comment above needs to be more clear I think}

% To encourage different certified regions to cover diverse areas of the C-space, we seed each region with a sample point, $s_{s}$ (typically known to be outside existing certified regions), and add a simple constraint that $s_{s}$ be contained in $\calP$. 

% Finally, we remark that due to the physical assumptions on the joints, we require that $\calP$ be entirely contained within the joint limit polyhedron $\calP_{\lim} = \{t \mid t_{l} \leq t \leq t_{u}\} = \{t \mid C_{\lim} t \leq q_{\lim}\}$. This requirement is captured by the following set of constraints \cite{freund1985complexity}:
% \begin{align}
%   \begin{aligned}
%   &~ Ct_{0} \leq d + \varepsilon
%   \\
%   & 
%   d_{i} + \varepsilon_{i} \mu_{i}^T q_{\lim} \geq 0,~\mu_{i} \geq 0  ~\forall~i \in \{1, \dots, m\} \\
%   &\calP_{\lim}^T\lambda_{i} = c_{i} ~\forall~i \in \{1, \dots, m\}
%   \end{aligned}
%   \label{E: eps limits}
% \end{align}

Combining the objective, the constraints of the maximal inscribed ellipsoid program, and the seed point condition with the program in \eqref{E: cert problem} yields the optimization program:
\begin{subequations}
\begin{gather}
    \max_{Q, s_0, C, d, \lambda, \nu, a_{\calA, \calB}, b_{\calA, \calB}}~ \logdet Q ~\subjectto 
    \\
    \norm{Qc_{i}}_{2} \leq d_{i} - c_{i}^Ts_{0} ~\forall~ i\in [m]
    \label{E: cert with ellipse1}\\
    Cs_{s} \leq d\label{E: cert with ellipse contain sample}\\ ~\norm{c_{i}}_{2}  \leq 1 ~\forall~ i \in [m]
    \label{E: cert with ellipse2}\\
    \alpha^{F, \calA_{j}}(a_{\calA, \calB}, b_{\calA, \calB}, s) = 
    \lambda_{0}^{\calA_{j},\calB}(s) + \sum_{i=1}^{m}\lambda_{i}^{\calA_{j}, \calB}(s)(d_{i} -c_{i}^Ts)
    ~\forall~ \text{pairs } (\calA, \calB)
    \label{E: cert with ellipse3}\\
    \beta^{F, \calB_{k}}(a_{\calA, \calB}, b_{\calA, \calB}, s) =
    \nu^{\calA, \calB_{k}}_{0}(s) + \sum_{i=1}^{m}\nu^{\calA, \calB_{k}}_{i}(s)(d_{i} -c_{i}^Ts)
    ~\forall~ \text{pairs } (\calA, \calB)
    \label{E: cert with ellipse4}\\
    Q \succeq 0,~\lambda(s), \nu(s) \in \bSigma.
    \end{gather}
    \label{E: cert with ellipse}
\end{subequations}
The condition $\calE_{\calP}\subset\calP$ is given by the constraint \eqref{E: cert with ellipse1}. \eqref{E: cert with ellipse contain sample} enforces that $\calP$ grows around $s_{s}$. \eqref{E: cert with ellipse3} and \eqref{E: cert with ellipse4} are the polynomial implication condition \eqref{E: alpha poly} \eqref{E: beta poly}: $s\in\calP=\{s \mid Cs\le d\}\implies \alpha\ge 0 \text{ and } \beta\ge 0$, proving the existence of the separating planes between the robot and the obstacles in the task space. The added constraint \eqref{E: cert with ellipse2} prevents numerically undesirable scaling of our polytope. While this program is attractive as a specification, it is not convex due to the bilinearity terms $Qc_i$ in \eqref{E: cert with ellipse1} and $\lambda_i d_i, \lambda_i c_i^T$ in \eqref{E: cert with ellipse3}\eqref{E: cert with ellipse4}. This bilinearity precludes simultaneous search of $\calP, \calE_{\calP}$ and the corresponding certificate.

However, the program \emph{is} convex when either the polytope $\calP$ is fixed or when the inscribed ellipse $\calE_{\calP}$ and the multiplier polynomials (save for $\lambda_{0}$ and $\nu_{0}$) are fixed. In the former case, given a polytope such that \eqref{E: cert with ellipse} is feasible, we simultaneously obtain a collision-free certificate $\calC_{\calP}$, maximal inscribed ellipsoid $\calE_{\calP}$, and an accompanying estimate of the size of $\calP$. In this situation, clearly \eqref{E: cert with ellipse2} becomes redundant. We remark that \eqref{E: cert with ellipse} is only marginally more complex than \eqref{E: cert problem} as it introduces only one additional semidefinite decision variable $Q$ and one vector decision variable $s_{0}$ with $m$ constraints which are not coupled to the multiplier polynomials.
\begin{figure}
    \centering
    \begin{tikzpicture}[scale=0.45, every node/.style={scale=0.7}]

%\draw[rotate = 30, opacity = 0.4]  (-1,1.5) ellipse (4 and 2.5);
\draw[opacity=0.2] (-2.414,-1.7495) -- (-5.1,-0.4);
\draw[opacity=0.2] (-3.714,-1.9495) -- (1.2796,0.6253);
\draw[opacity=0.2] (-3.9825,1.4686) -- (0.4,3.3);
\draw[opacity=0.2] (-3.1516,2.0658) -- (-5.1,-1.2);
\draw[opacity=0.2] (-0.7355,3.6263) -- (2.2,1.5);
\draw[opacity=0.2] (1.7,2.7) -- (-0.1456,-0.9485);
\draw[thick, opacity = 0.2] (-4.7325,-0.5675) -- (-2.8677,-1.4952) -- (0.4168,0.1389) -- (1.413,2.0447) -- (0,3.1) -- (-3.2915,1.738)--cycle;
\draw[rotate = 30, opacity = 0.2 ]  (-1,1.5) ellipse (3 and 1.5);
\draw (-5.2914,3.146) -- (1.6,4.5);
\draw (-4.7,3.5) -- (-5.7,-2.2);
\draw (0.6,4.6) -- (3.1,2.8);
\draw (3.1,3.3) -- (1.4981,-0.9552);
\draw (2.1,0) -- (-2.6,-3);
\draw (-6.2,-1.6) -- (-1.792,-3.019);
\draw[<->] (-1.1324,3.9542) -- (-0.8531,2.7303);
\draw[<->] (1.7,3.8) -- (0.7239,2.5926);

\draw[<->] (-5.2982,0.086) -- (-4.3254,-0.071);
\draw[<->] (0.9986,1.2045) -- (2.0972,0.6987);
\draw[<->] (-3.728,-1.0843) -- (-4.1632,-2.2531);
%\node at (-0.4116,3.0957) {$\delta_i$};
\begin{scope}[shift = {(-10, 0)}]
\draw[EDB] (-5.3,4.8) -- (-4.9893,4.8767) -- (-4.0159,5.042) -- (-2.7487,5.2073) -- (-1.6651,5.244) -- (-0.9305,5.2257) -- (-0.0122,4.9686) -- (1.1815,4.5462) -- (2.1733,4.4543) -- (2.6,4.5) -- (2.6,6.4) -- (-5.3,6.4)--cycle;
\draw[very thick, opacity = 0.8]  plot[smooth, tension=.7] coordinates {(-5.3,4.8) (-1.4447,5.2808) (0.9243,4.6196) (1.9529,4.436) (2.5773,4.5462)};

\draw[opacity=0.4] (-2.414,-1.7495) -- (-5.1,-0.4);
\draw[opacity=0.4] (-3.714,-1.9495) -- (1.2796,0.6253);
\draw[opacity=0.4] (-3.9825,1.4686) -- (0.4,3.3);
\draw[opacity=0.4] (-3.1516,2.0658) -- (-5.1,-1.2);
\draw[opacity=0.4] (-0.7355,3.6263) -- (2.2,1.5);
\draw[opacity=0.4] (1.7,2.7) -- (-0.1456,-0.9485);
\draw[thick] (-4.7325,-0.5675) -- (-2.8677,-1.4952) -- (0.4168,0.1389) -- (1.413,2.0447) -- (0,3.1) -- (-3.2915,1.738)--cycle;
\draw[rotate = 30 ]  (-1,1.5) ellipse (3 and 1.5);
\end{scope}

\begin{scope}[shift = {(10, 0)}]
\draw[EDB] (-5.3,4.8) -- (-4.9893,4.8767) -- (-4.0159,5.042) -- (-2.7487,5.2073) -- (-1.6651,5.244) -- (-0.9305,5.2257) -- (-0.0122,4.9686) -- (1.1815,4.5462) -- (2.1733,4.4543) -- (2.6,4.5) -- (2.6,6.4) -- (-5.3,6.4)--cycle;
\draw[very thick, opacity = 0.8]  plot[smooth, tension=.7] coordinates {(-5.3,4.8) (-1.4447,5.2808) (0.9243,4.6196) (1.9529,4.436) (2.5773,4.5462)};

\draw[opacity=0.2] (-5.2914,3.146) -- (1.9322,4.5155);
\draw[opacity=0.2] (-4.7,3.5) -- (-5.7,-2.2);
\draw[opacity=0.2] (0.8,4.5) -- (3.1,2.8);
\draw[opacity=0.2] (3.1,3.3) -- (1.4981,-0.9552);
\draw[opacity=0.2] (2.1,0) -- (-2.6,-3);
\draw[opacity=0.2] (-6.2,-1.6) -- (-1.792,-3.019);
\draw[rotate = 33 ]  (-0.7,1.5) ellipse (4.4 and 2.6);
\end{scope}

\draw[<->] (-0.8123,-0.4644) -- (-0.1977,-1.4495);
\node at (-0.8335,2.3025) {$\delta_1$};

\node at (0.2036,2.2159) {$\delta_6$};
\node at (0.355,1.3321) {$\delta_5$};
\node at (-1.1978,0.1527) {$\delta_4$};
\node at (-3.2325,-0.4453) {$\delta_3$};
\node at (-3.698,0.0924) {$\delta_2$};
\draw[EDB] (-5.3,4.8) -- (-4.9893,4.8767) -- (-4.0159,5.042) -- (-2.7487,5.2073) -- (-1.6651,5.244) -- (-0.9305,5.2257) -- (-0.0122,4.9686) -- (1.1815,4.5462) -- (2.1733,4.4543) -- (2.6,4.5) -- (2.6,6.4) -- (-5.3,6.4)--cycle;
\draw[very thick, opacity = 0.8]  plot[smooth, tension=.7] coordinates {(-5.3,4.8) (-1.4447,5.2808) (0.9243,4.6196) (1.9529,4.436) (2.5773,4.5462)};

\draw[thick] (5.2332,3.2523) -- (11.0376,4.3368) -- (12.931,2.9145) -- (11.7665,-0.2144) -- (7.6332,-2.8278) -- (4.371,-1.7965)--cycle;

\end{tikzpicture}
    \caption{In \eqref{E: cert with polytope} we search for the maximum amount the polytopes faces can be pushed away from the current inscribed ellipse without violating the certificate found in the previous step.
    % In \eqref{E: cert with polytope} we search for the maximal pushback of the faces of the polytope away from the inscribed ellipse without violating the certificate found in the previous step.
    }
    \label{F: push back}
    \vspace{-13.5pt}
\end{figure}
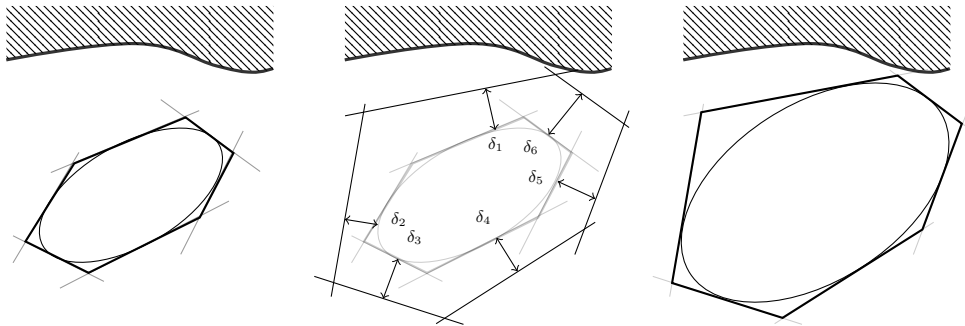

Conversely, from a solution of \eqref{E: cert with ellipse} we are also capable of searching for a new collision-free polytope with an accompanying certificate. To ensure growth of the polytopic region, our objective is to push the faces of the current polytope away from the inscribed ellipsoid without violating our certificate as in Fig \ref{F: push back}. To accomplish this, we present the following minor modification of \eqref{E: cert with ellipse}
\begin{gather}
    \begin{gathered}
    \max_{C, d, \delta, \lambda_{0}, \nu_{0}, a_{\calA, \calB}, b_{\calA, \calB}}~ \prod_{i=1}^{m} (\delta_{i} + \varepsilon_{0}) ~\subjectto 
    \\
    \norm{Qc_{i}}_{2} \leq d_{i} - \delta_{i} - c_{i}^Ts_{0}, ~ \delta_{i} \geq 0 ~\forall~ i\in [m]
    \\
    \eqref{E: cert with ellipse contain sample}, \eqref{E: cert with ellipse2},
    \eqref{E: cert with ellipse3}, \eqref{E: cert with ellipse4}\\
    \lambda_{0}(s), \nu_{0}(s) \in \bSigma
    \end{gathered}
    \label{E: cert with polytope}
\end{gather} 
where $\varepsilon_{0} > 0$ is some positive constant ensuring that the objective is never $0$. We emphasize that we have fixed the multipliers $\lambda_{i}$ and $\nu_{i}$ for $i > 0$ and instead search for $d_{i}$ and $c_{i}$.

In this program, we introduce the variable $\delta_{i}$ which measures the distance between the hyperplane $c_i^Ts=d_i$ and the inscribed ellipse. Searching over the variable $C$ allows the normal of the polytope to rotate in order to accomplish a greater change in margin. The remaining constraints enforce that the polytope continue to contain the ellipse $\calE_{\calP}$ while also generating a new set of separating planes within the collision-free certificate $\calC_{\calP}$. The alternation procedure is outlined in Algorithm \ref{A: Bilinear Alternation} and can be seen as a generalization of the {\sc{Iris}} algorithm from \cite{deits2015computing}. Notice that the volume of the inscribed ellipsoid $\calE_{\calP}$ is guaranteed to increase in each iteration of Algorithm \ref{A: Bilinear Alternation}.

\begin{algorithm}
\caption{
Given an initial polytopic region $\calP_{0}$ for which \eqref{E: cert with ellipse} is feasible, return a new polytopic region $\calP_{i}$ with a maximal inscribed ellipse $\calE_{\calP_{i}}$ with larger volume than $\calE_{\calP_{0}}$ and a collision-free certificate $\calC_{\calP_{i}}$. Notice that since $\calP_{0}$ is assumed collision-free, programs \eqref{E: cert with ellipse} and \eqref{E: cert with polytope} are always feasible. \eqref{E: cert with ellipse} and \eqref{E: cert with polytope} are alternated until the change in the volume of the ellipse $\calE_{\calP}$ is less than some $tolerance$.
}\label{A: Bilinear Alternation}
\SetAlgoLined
 \LinesNumbered
  \SetKwRepeat{Do}{do}{while}
 $i \gets 0$
 \\
 \Do{$\left(\textbf{vol}(\calE_{\calP_{i}}) - \textbf{vol}(\calE_{\calP_{i-1}})\right) / \textbf{vol}(\calE_{\calP_{i-1}})  \geq$ tolerance}{
 $(\calE_{\calP_{i}}, \calC_{\calP_{i}}) \gets$ Solution of \eqref{E: cert with ellipse} with data $(\calP_{i})$
 \\
 $(\calP_{i+1}, \calC_{\calP_{i+1}}) \gets$ Solution of \eqref{E: cert with polytope} with data $(\calE_{\calP_{i}}, \calC_{\calP_{i}})$
 \\
 $i \gets i+1$
 }
 \Return $(\calP_{i}, \calC_{\calP_{i}})$
\end{algorithm}

\begin{remark}
To make Algorithm \ref{A: Bilinear Alternation} numerically tractable, in Appendix \ref{S: Practical}, we discuss practical aspects to scale the algorithm to realistic examples. 
\end{remark}
\section{Results} \label{S: Results}
In this section, we demonstrate the use of Algorithm \ref{A: Bilinear Alternation} on three representative examples.
% we describe some results of running Algorithm \ref{A: Bilinear Alternation} on three representative examples.
In the first example, we consider a 3-DOF system for which we can visualize the entire configuration space. This enables us to plot the resulting regions of our algorithm. Next, we analyze the typical run times on the 7-DOF KUKA iiwa. Finally, we demonstrate the scalability of our algorithm on a 12-DOF system composed of two KUKA iiwas (with the wrist joint fixed as the wrist  link is symmetric about the joint axis). All convex programs are solved using Mosek v9.2 \cite{mosek} running on Ubuntu 20.04 with an Intel 11th generation i9 processor. Code is publicly available on \href{https://github.com/AlexandreAmice/drake/tree/C_Iris}{Github}\footnote{\url{https://github.com/AlexandreAmice/drake/tree/C_Iris}}.

% \red{
% \begin{itemize}
%     \item Show table with runtimes. Compare to IBEX and how IBEX can't even run on the simple 3-DOF example
%     \item Volume growth plots (Hongkai)
%     \begin{itemize}
%         \item Confined growth
%         \item Growth in free space
%     \end{itemize}
%     \item Show iiwa configurations in certified regions that are close to collision boundary
% \end{itemize}
% }
 
\subsection{3-DOF Flipper System} \label{S: Pinball}
In this example, we consider the 3-DOF system in the left-most panel of Fig \ref{F: collision constraint pinball} consisting of two iiwas with all joints save those at the ends of the orange links fused. This system has few degrees of freedom, but maintains rich, realistic collision geometries. As the C-space is three-dimensional, we are able to visualize the collision constraint by using marching cubes \cite{lorensen1987marching}. This is plotted as the red mesh in the right two panels of Fig \ref{F: collision constraint pinball}.

We run Algorithm \ref{A: Bilinear Alternation} on 11 different regions initialized according to Algorithm \ref{A: SNOPT IRIS} described in Appendix \ref{S: Seeding}. Our approach is able to grow regions whose union describes 90\% of C-free. In Fig \ref{F: collision constraint pinball}, we visualize a subset of these 11 regions demonstrating that each region covers a substantial amount of the C-free on their own. In step 3 of Algorithm \ref{A: Bilinear Alternation}, the largest SOS takes 0.008s to solve, and in step 4 of Algorithm \ref{A: Bilinear Alternation}, the SOS takes 0.647s to solve.
 \begin{figure*}
    \centering
    \begin{subfigure}[t]{0.32\textwidth}
    \centering
        \includegraphics[width = 0.98\textwidth,trim={8cm 12cm 8cm 4cm},clip]{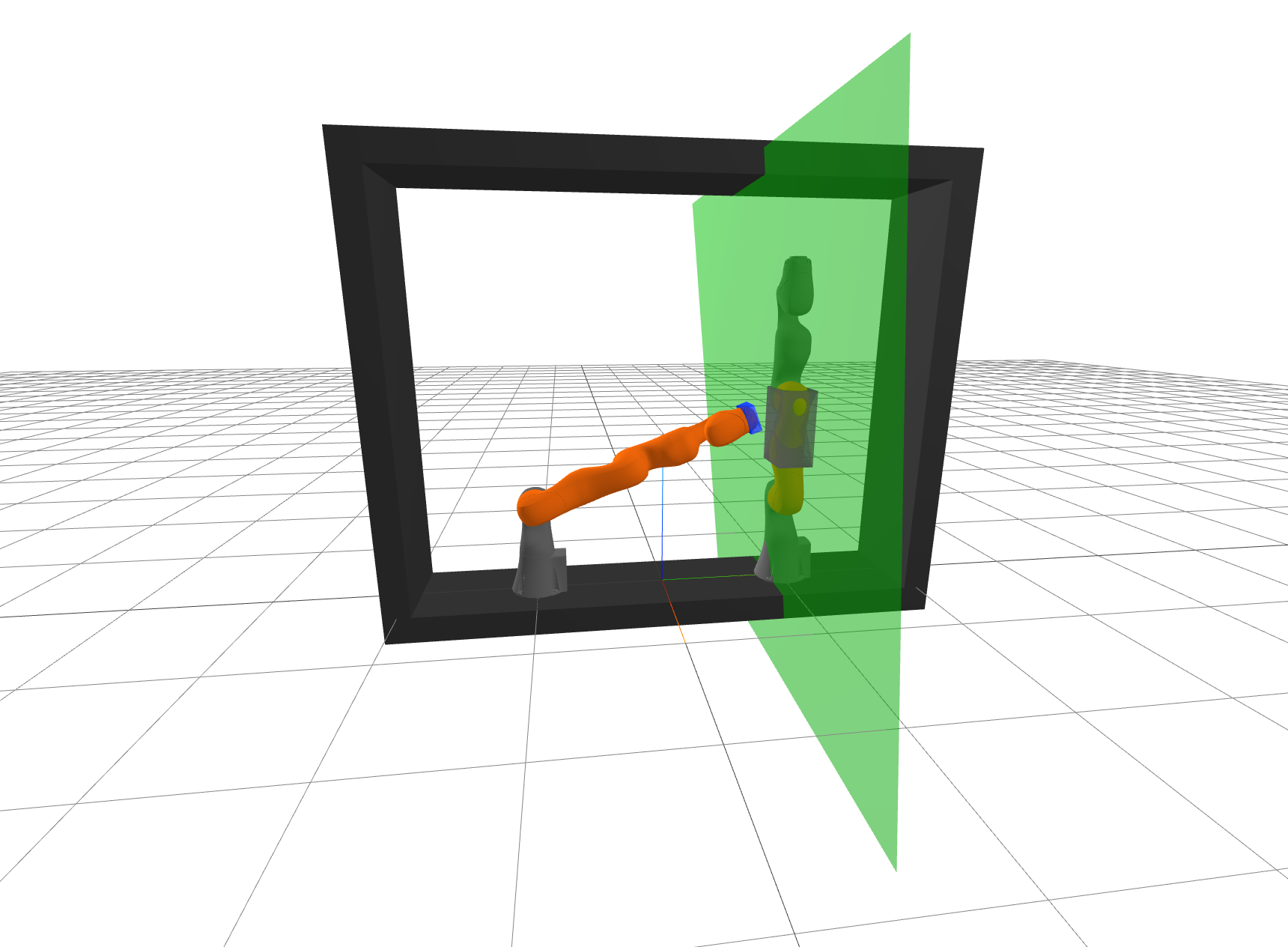}
        %\caption{Start of Trajectory} \label{F: cert cfg2}
    \end{subfigure}
    \begin{subfigure}[t]{0.32\textwidth}
    \centering
        \includegraphics[width = 0.98\textwidth, trim={8cm 12cm 8cm 4cm},clip]{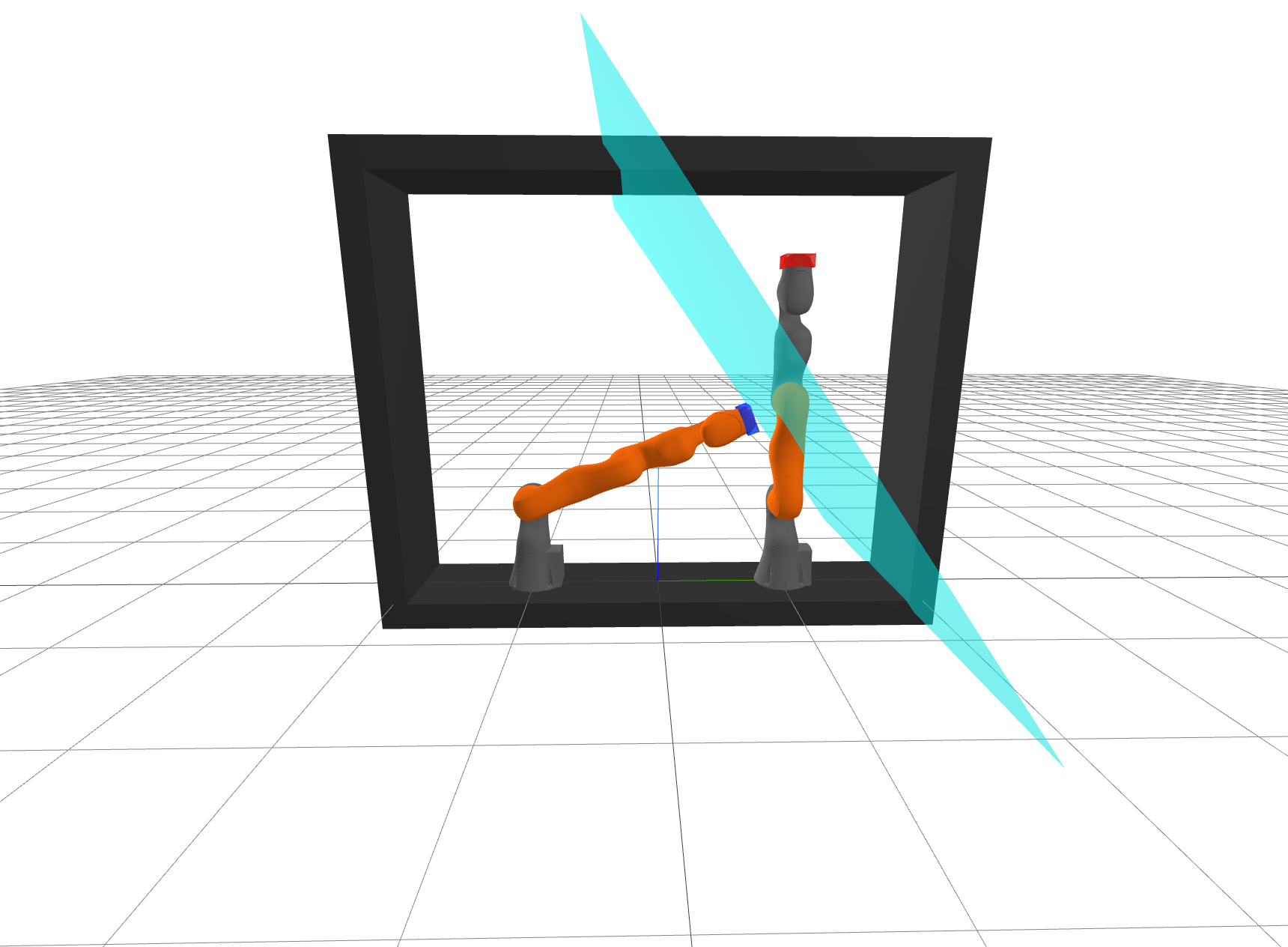}
        %\caption{Middle of Trajectory} \label{F: cert cfg1}
        \vspace{ 0.3 cm}
    \end{subfigure}
    \begin{subfigure}[t]{0.32\textwidth}
    \centering
        \includegraphics[width = 0.98\textwidth, trim={6cm 4cm 10cm 12cm},clip]{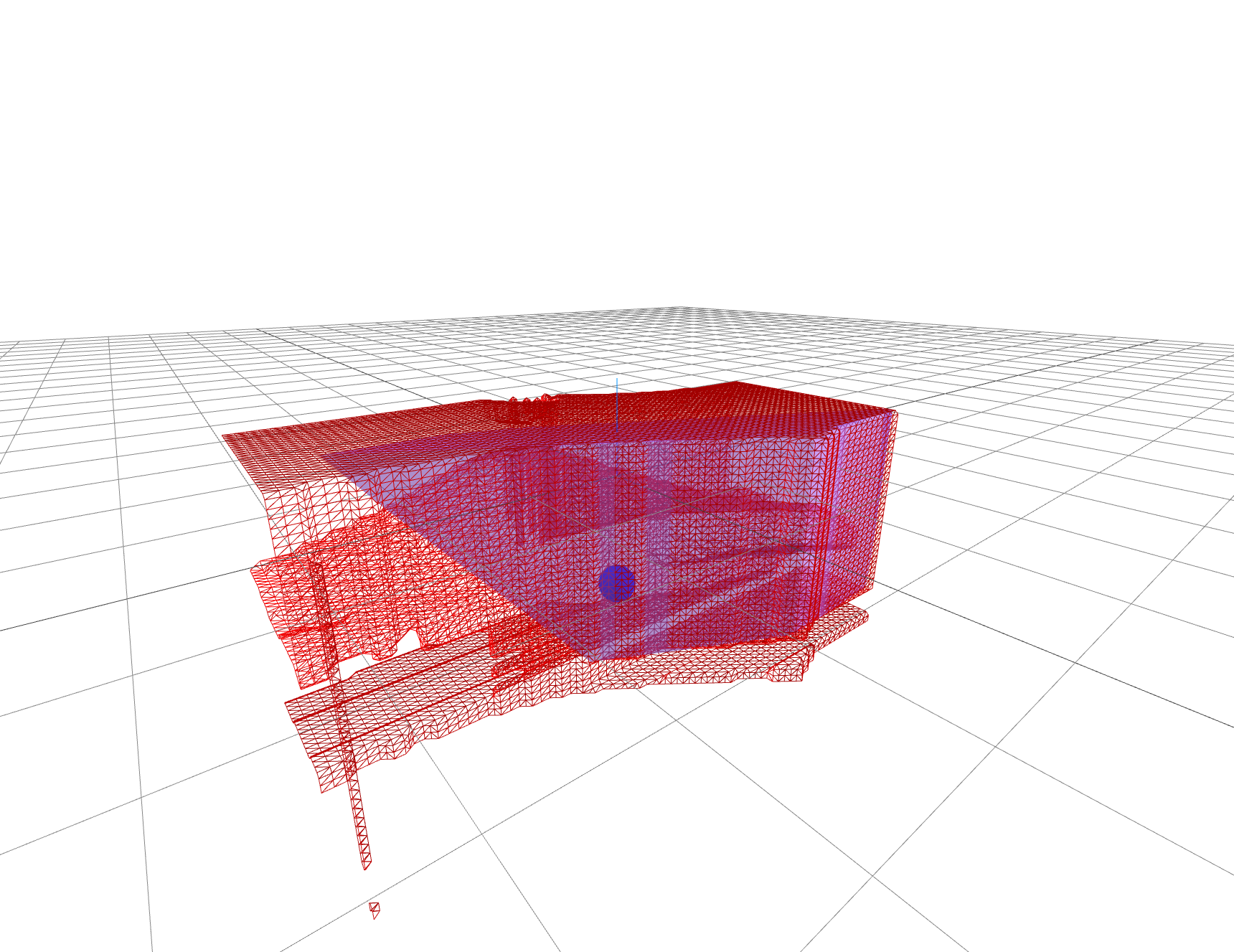}
        %\caption{Start of Trajectory} \label{F: cert cfg2}
    \end{subfigure}
    \begin{subfigure}[t]{0.32\textwidth}
    \centering
        \includegraphics[width = 0.98\textwidth, trim={8cm 11cm 8cm 5cm},clip]{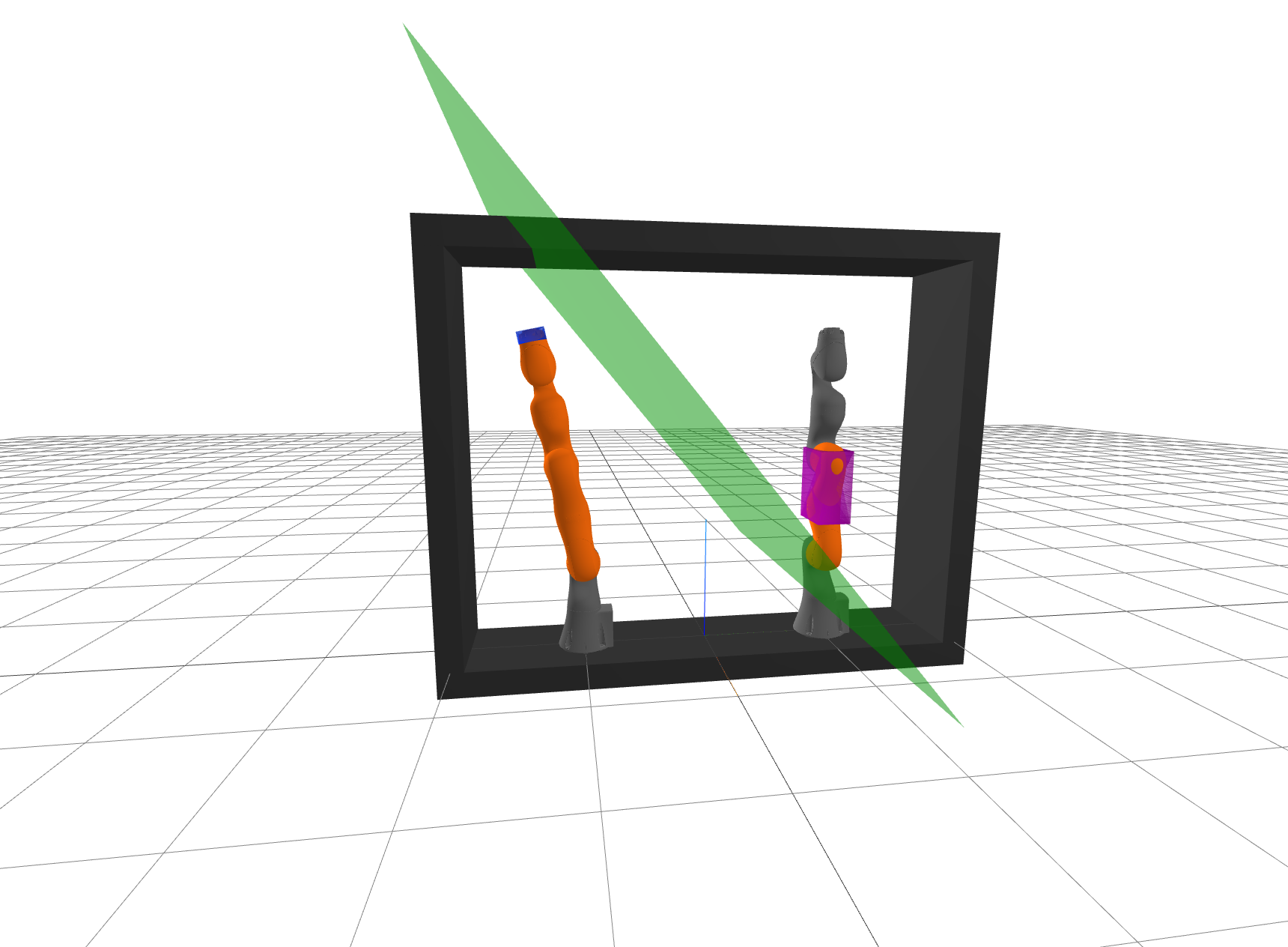}
        %\caption{Middle of Trajectory} \label{F: cert cfg1}
        \vspace{ 0.3 cm}
    \end{subfigure}
    \begin{subfigure}[t]{0.32\textwidth}
    \centering
        \includegraphics[width = 0.98\textwidth, trim={8cm 11cm 8cm 5cm},clip]{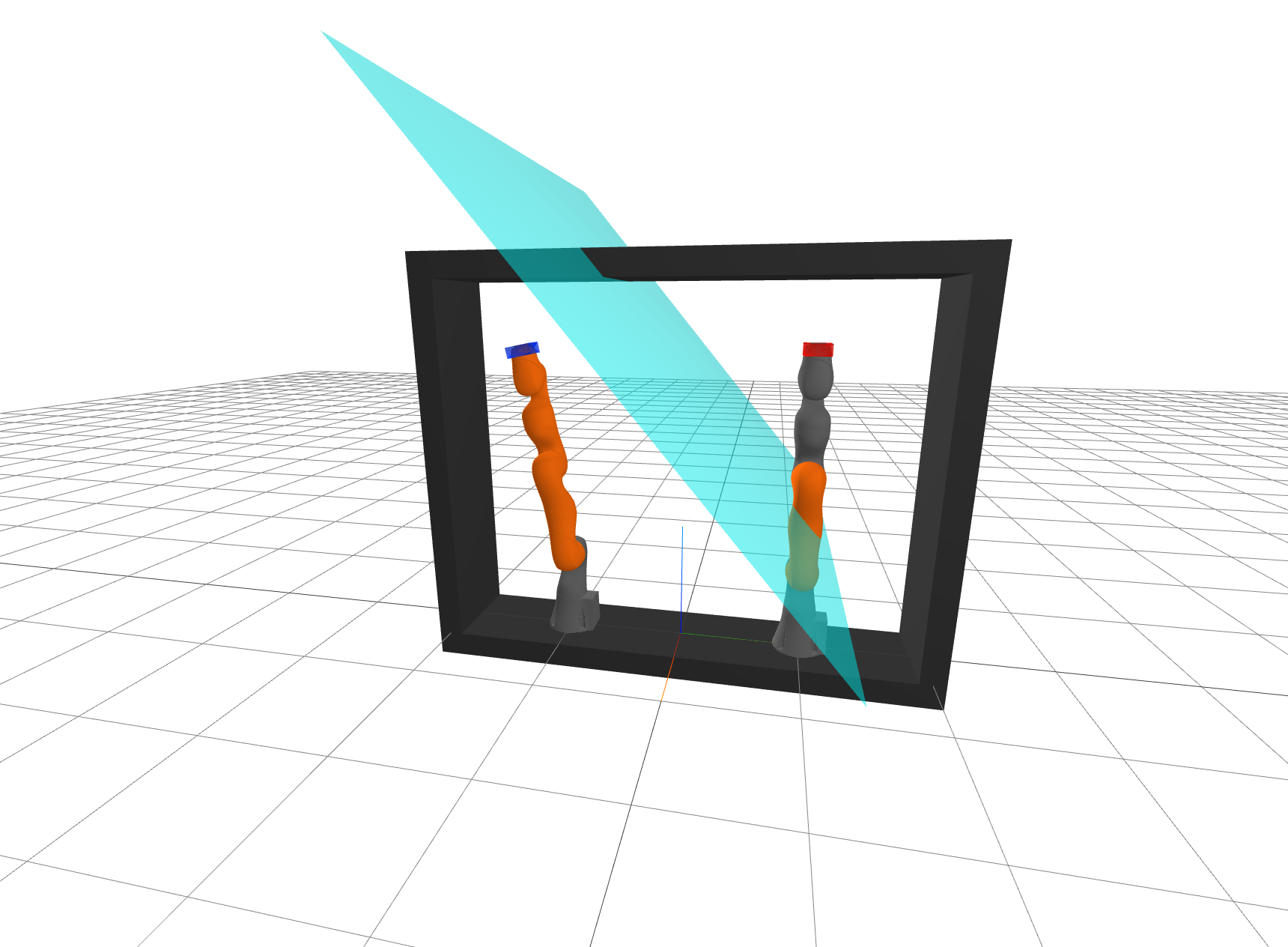}
        %\caption{Start of Trajectory} \label{F: cert cfg2}
    \end{subfigure}
    \begin{subfigure}[t]{0.32\textwidth}
    \centering
        \includegraphics[width = 0.98\textwidth, trim={6cm 4cm 10cm 12cm},clip]{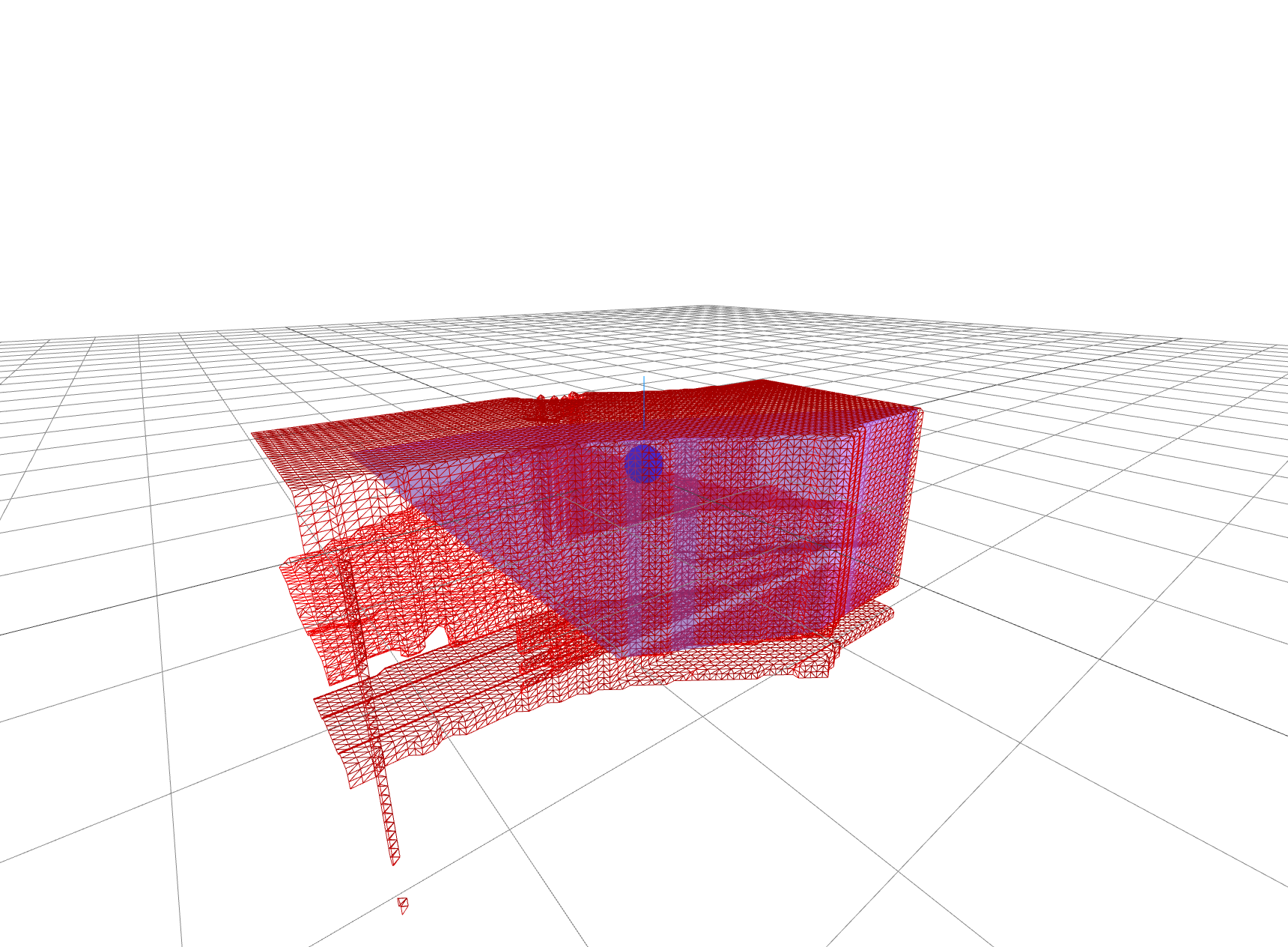}
        %\caption{Middle of Trajectory} \label{F: cert cfg1}
    \end{subfigure}
    \caption{\small{Algorithm \ref{A: Bilinear Alternation} generates a set of separating planes between each collision body that vary as the configuration moves (denoted by the blue sphere in the right most pane). As the configuration varies in the purple C-free region, the separating planes in green and teal can move to accommodate the relative shifts in the robot positions.}}
    \label{F: plane certificates}
    \vspace{-13.5pt}
\end{figure*}

 In Fig \ref{F: plane certificates}, we demonstrate the certification of a single, particularly large region.  The green and teal plane certify that the end-effector of the left arm (highlighted in blue) does not collide with the middle link (highlighted in purple) and the top link (highlighted in red) respectively for all configurations in the purple region shown in right hand panes of \ref{F: plane certificates}.

\noindent

\subsection{7-DOF KUKA with Shelf}
\begin{figure*}
    \centering
    \begin{subfigure}[t]{0.32\textwidth}
    \centering
        \includegraphics[width = 1.0\textwidth, trim={18cm 8cm 22cm 12cm},clip]{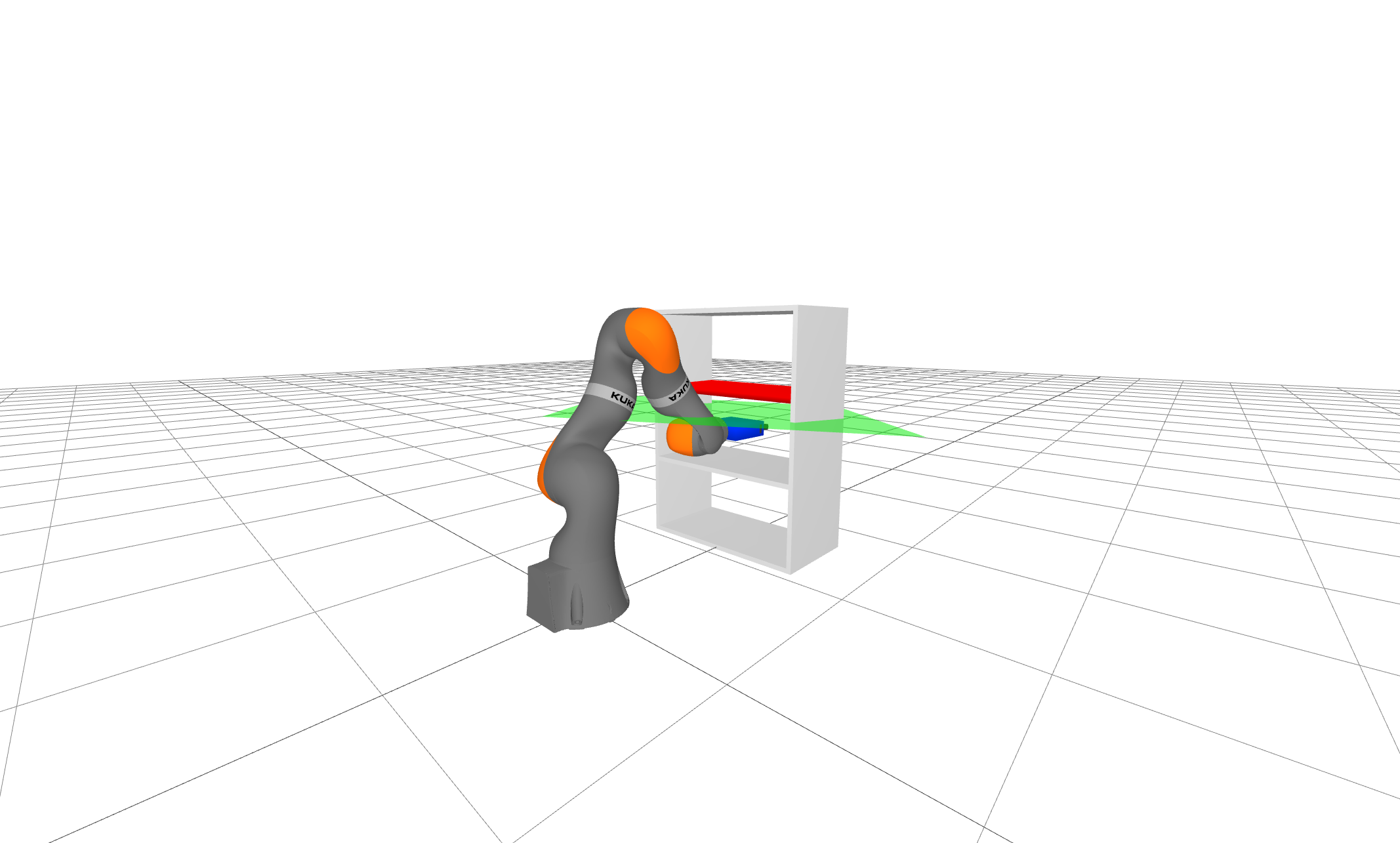}
        %\caption{Start of Trajectory} \label{F: cert cfg2}
    \end{subfigure}
    \begin{subfigure}[t]{0.32\textwidth}
    \centering
        \includegraphics[width = 1.0\textwidth, trim={18cm 8cm 22cm 12cm},clip]{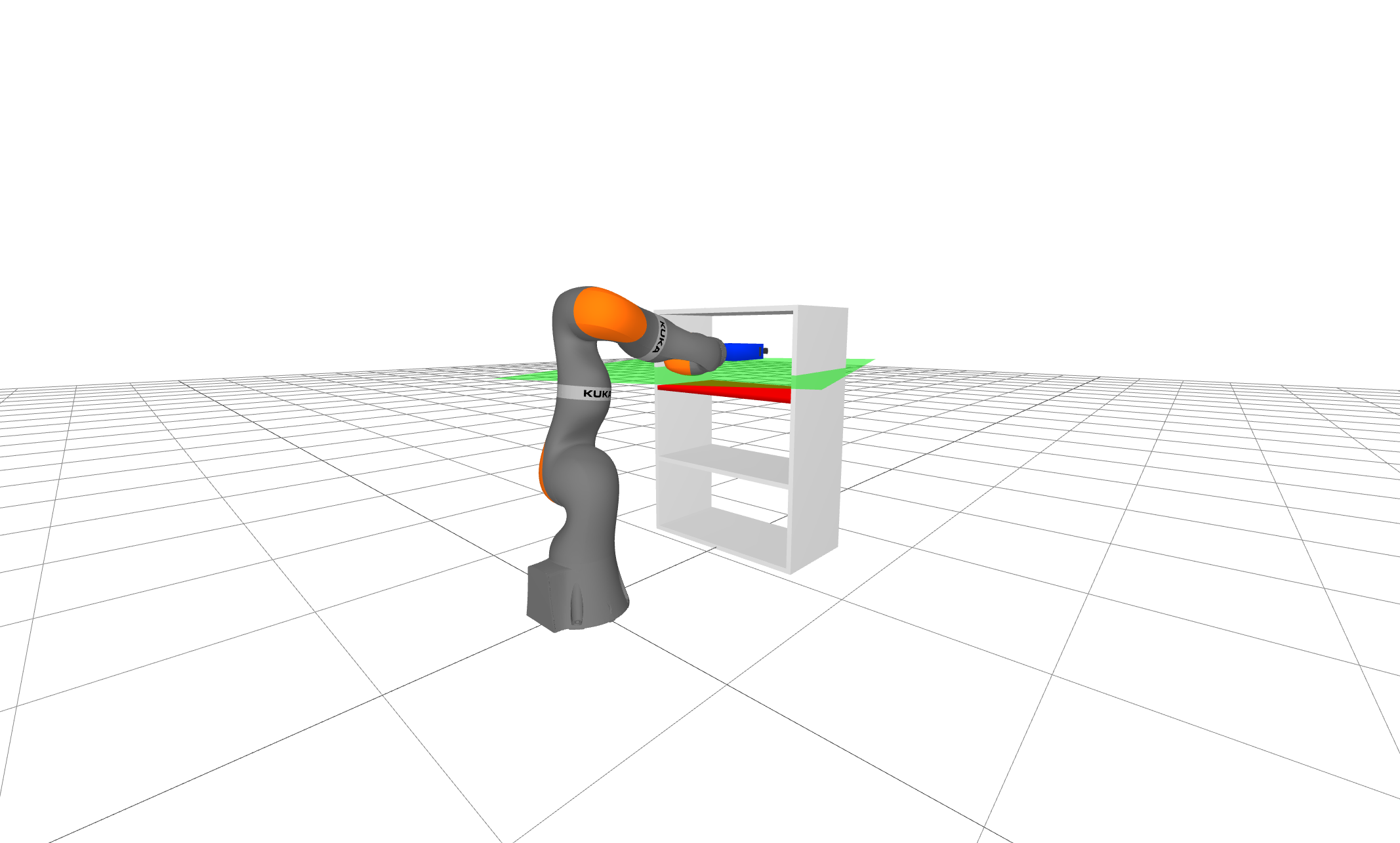}
        %\caption{Middle of Trajectory} \label{F: cert cfg1}
        \vspace{ 0.3 cm}
    \end{subfigure}
    \begin{subfigure}[t]{0.32\textwidth}
    \centering
        \includegraphics[width = 1.0\textwidth, trim={18cm 8cm 22cm 12cm},clip]{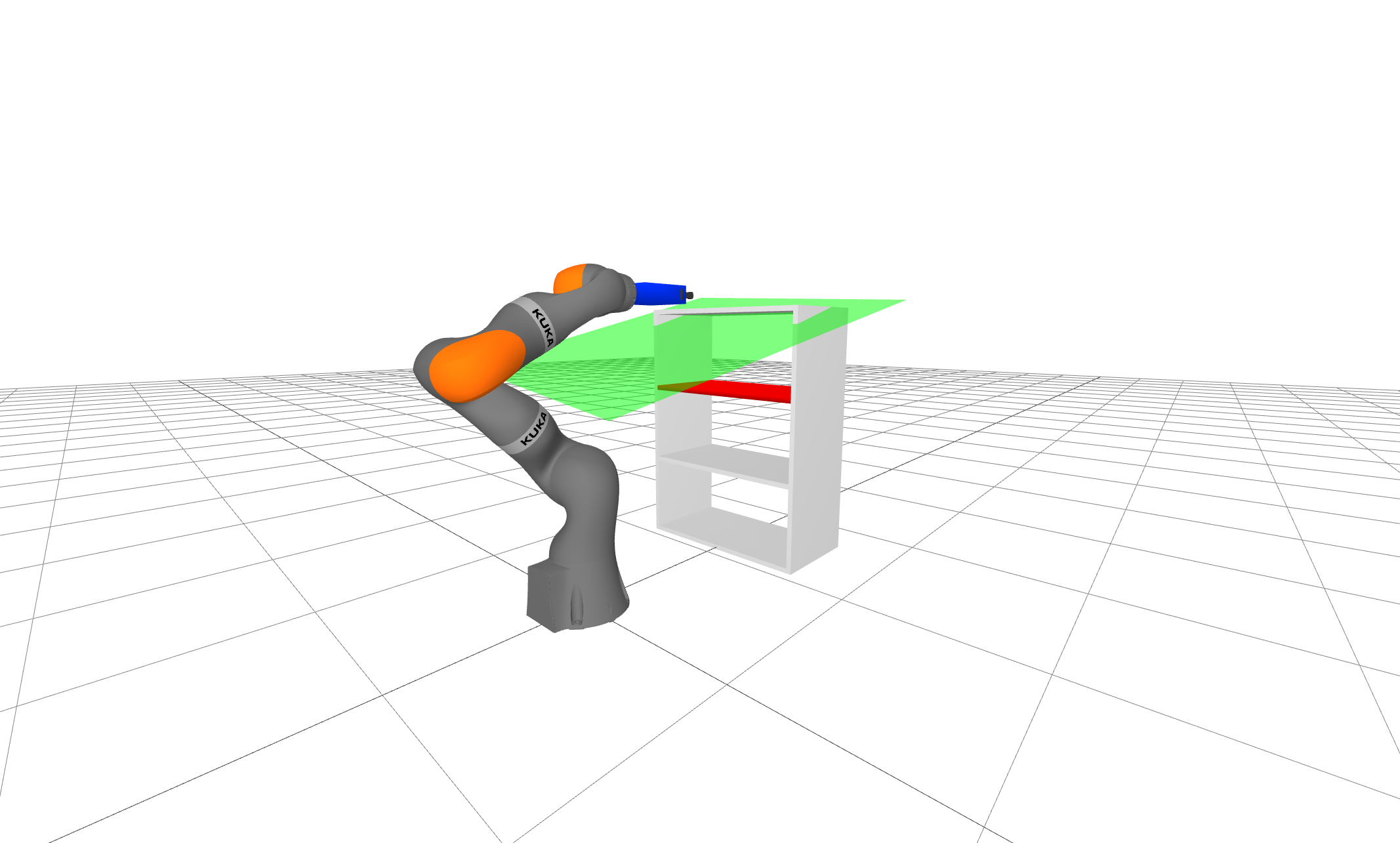}
        %\caption{Start of Trajectory} \label{F: cert cfg2}
    \end{subfigure}
    \caption{\small{7-DOF iiwa  example. We highlight one pair of collision geometries (blue on robot gripper and red on the shelf), together with their separating plane (green).}}
    \label{F: collision constraint}
    \vspace{-13.5pt}
\end{figure*}

\begin{figure}
    \centering
    \begin{subfigure}{0.45\textwidth}
    \centering
    \includegraphics[width=0.7\textwidth]{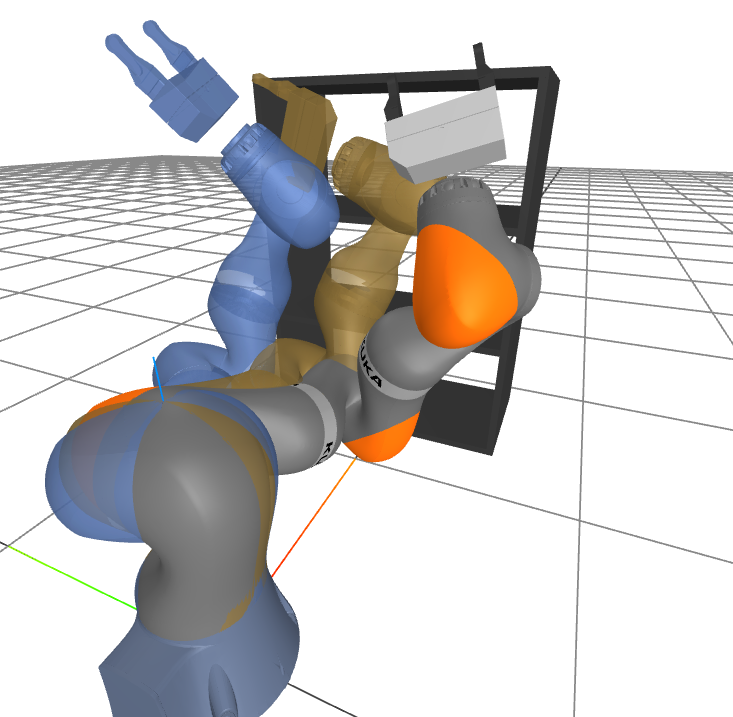}
    \end{subfigure}
    \begin{subfigure}{0.45\textwidth}
    \centering
    \includegraphics[width=0.96\textwidth]{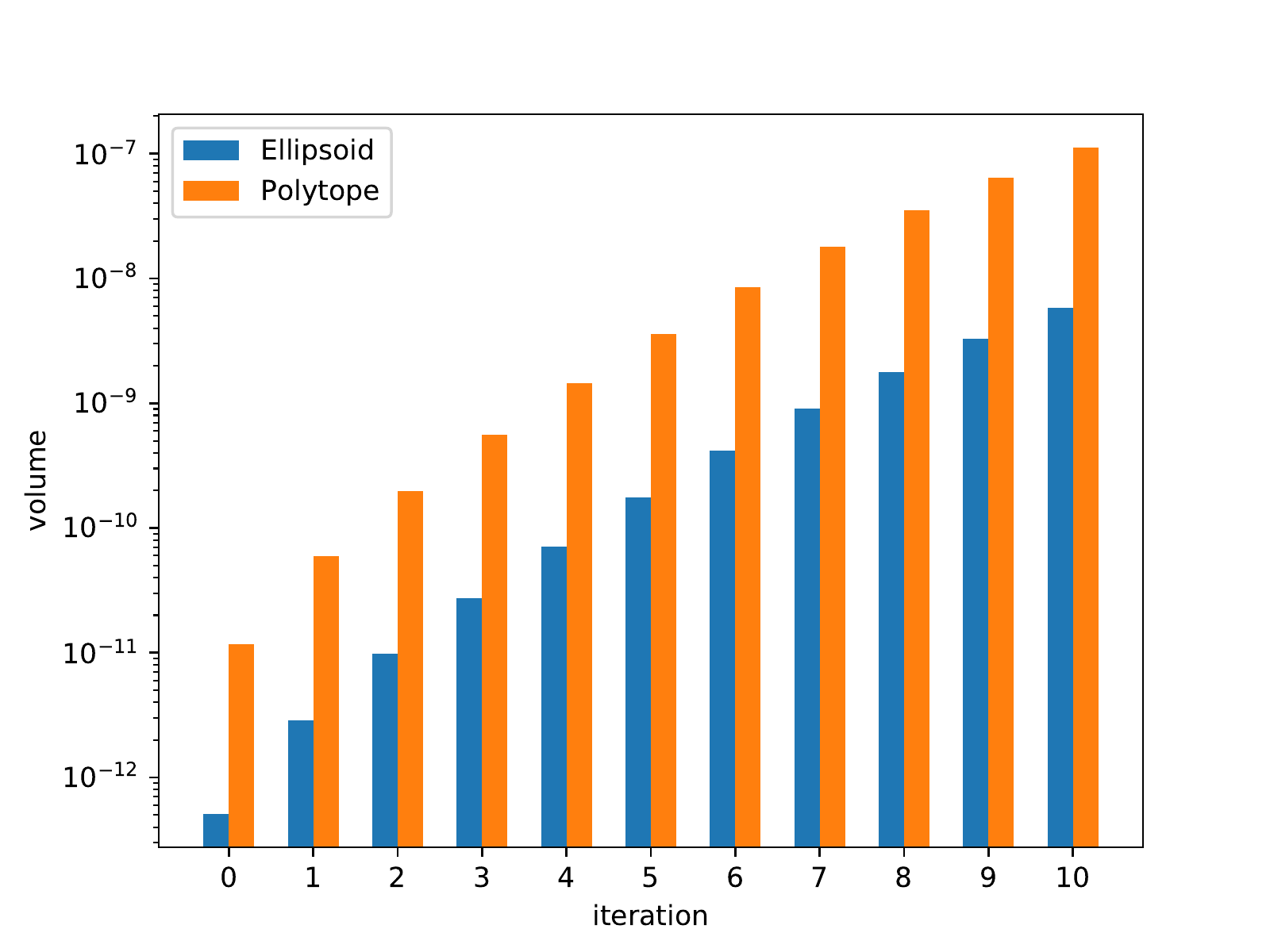}
    \end{subfigure}
    \caption{Left: multiple sampled postures (in different colors) in a certified C-free region. Right: The growth of volume with Algorithm \ref{A: Bilinear Alternation} in each iteration.}
    \label{fig:iiwa_shelf_volume}
    \vspace{-13.5pt}
\end{figure}

\vspace{-3pt}
We apply our approach to a 7-DOF KUKA iiwa arm to certify and search for C-free regions with a shelf, as shown in Fig \ref{F: collision constraint}. We also show the growth of volume for one certified C-free region in Fig \ref{fig:iiwa_shelf_volume}. The volume increases by a factor of 10,000 in 11 iterations of Algorithm \ref{A: Bilinear Alternation}, and covers a range of configurations (Fig \ref{fig:iiwa_shelf_volume} left). In step 3 of Algorithm \ref{A: Bilinear Alternation}, the largest SOS take 54s to solve, and in step 4 of Algorithm \ref{A: Bilinear Alternation}, the SOS takes 8s to solve.


\subsection{12-DOF Bimanual Example}
Finally we demonstrate the scalability of our approach on a 12-DOF system of two KUKA iiwa arms (with fixed wrist joint) to find C-free regions avoiding self-collision. 
In Fig \ref{Fig: dual_iiwa}, we visualize several postures in one certified C-free region. At one sampled posture, the closest distance between the collision geometries is 7.3mm (See Fig \ref{Fig: dual_iiwa} right), indicating our certified C-free region is extremely tight. 12-DOF is quite high-dimensional for any sampling-based algorithm, and the solution times for generating our strong deterministic certificate increases significantly, too, compared to the single arm case. Solving the largest SOS program in line 3 of Algorithm \ref{A: Bilinear Alternation} takes 105 minutes, and the SOS program in line 4 of Algorithm \ref{A: Bilinear Alternation} take 4 minutes.

\section{Conclusion and Future Work}
In this work, we present an approach to find large certifiable C-free regions for robot manipulators. Our approach certifies a polytopic region in the tangent-configuration space to be collision-free through convex optimization. Moreover, we give a practical, iterative algorithm for finding and growing these C-free regions. Our method works in arbitrary dimensions and scales to reasonable, realistic examples in robotic manipulation. Such certified regions find practical use in both randomized and optimization-based collision-free motion planning algorithms.

While this paper assumed a system composed of only revolute joints and collision bodies defined as V-rep polytopes, we believe that the essential machinery can be applied to arbitrary algebraic joints and convex bodies. We intend to investigate such extensions in future work. Additionally, in Appendix \ref{S: iris pseudocode} we have also given a non-linear optimization algorithm for very rapidly proposing regions when the cost of many alternations becomes prohibitive. Future work will explore a tighter integration of Algorithms \ref{A: Bilinear Alternation} and \ref{A: SNOPT IRIS}.

% Such regions find practical use in planning algorithms such as the one given in our companion paper \cite{marcucci2022convex}. In future work, we plan to integrate these two papers into a single, global planning method. Additionally, while this paper assumed a plant composed of only revolute joints and collision bodies defined as V-rep polytopes, the essential machinery can be applied to arbitrary algebraic joints and convex bodies. We intend to investigate these extensions.

\putbib
\end{bibunit}

\newpage

\appendix
\begin{bibunit}
\section{Definition of Archimedean}\label{A: Archimedean}
In this section we formally define the Archimedean property that appears in Theorem \ref{T: Putinar}. 
\begin{definition} \label{D: Archimedean}
A semialgebraic set $\calS_{g} = \{x \mid g_{i}(x) \geq 0, i \in [n]\}$ is Archimedean if there exists $N \in \setN$ and $\lambda_{i}(x) \in \bSigma$ such that:
\begin{align*}
    N - \sum_{i=1}^{n}x_{i}^2 = \lambda_{0}(x) + \sum_{i=1}^{n} \lambda_{i}(x)g_i(x)
\end{align*}
\end{definition}

\section{Practical Aspects} \label{S: Practical}
 In this section we discuss important ways to scale Algorithm \ref{A: Bilinear Alternation}. In sections \ref{S: Frame Selection} and \ref{S: Basis Selection} we describe design choices which dramatically reduce the size of the SOS programs used in the alternations. Next, we discuss aspects of Algorithm \ref{A: Bilinear Alternation} which can be  parallelized, reducing solve time in the alternations. Finally, we describe an extension to the original {\sc{Iris}} algorithm \cite{deits2015computing} which can be used to rapidly propose a very large region that is likely, but not certified, to be collision-free. Seeding Algorithm \ref{A: Bilinear Alternation} with such a region can dramatically reduce the number of iterations required to obtain a satisfyingly large volume.

 %In this section, we discuss important practical aspects of implementing Algorithm \ref{A: Bilinear Alternation} that enable us to scale it to realistic, practical problems. In Section \ref{S: Frame Selection} and \ref{S: Basis Selection} we describe the optimal choice of frame in which to express the forward kinematics and bases in which to express the multiplier polynomials. This has dramatic effect on the size of the semidefinite variables used to solve the programs in the alternations. Next, we remark on aspects of Algorithm \ref{A: Bilinear Alternation} which can be trivially parallelized. We conclude with a note on seeding the algorithm with an initial, feasible polytope.

\subsection{Choosing the Reference Frame}  \label{S: Frame Selection}
The polynomial implications upon which the programs \eqref{E: cert with ellipse} and \eqref{E: cert with polytope} are based require choosing a coordinate frame between each collision pair $\calA$ and $\calB$. However, as the collision-free certificate between two different collision pairs can be computed independently of each other, we are free to choose a different coordinate frame to express the kinematics for each collision pair. This is important in light of \eqref{E: gen trig poly} and \eqref{E: rational sub} that indicate that the degree of the polynomials $\leftidx{^F}f^{\calA_{j}}$  and $\leftidx{^F}g^{\calA_{j}}$ are equal to two times the number of joints lying on the kinematic chain between frame $F$ and the frame for $\calA$. For example, the tangent-configuration space polynomial in the variable $s$ describing the position of the end-effector of a 7-DOF robot is of total degree $14$ when written in the coordinate frame of the robot base. However, when written in the frame of the third link, the polynomial describing the position of the end effector is only of total degree $(7-3)\times 2=8$. This observation is also used in  \cite{trutman2020globally} to reduce the size of the optimization program. 

%Fortunately, the polynomials implications of \eqref{E: cert with ellipse} and \eqref{E: cert with polytope} are decoupled in the sense that the choice of reference frame $F$ for the forward kinematics can be chosen independently for each collision pair $(\calA, \calB)$. While choosing a reference frame $F$ near $\calA$ reduces the degree of the polynomial needed to express the forward kinematics of $\calA$, it comes at the cost of increasing the degree of the polynomial needed to express the kinematics of $\calB$. As we shall see in Section \ref{S: Basis Selection}, the size of the semidefinite variables in \eqref{E: cert with ellipse} and \eqref{E: cert with polytope} scale as the square of the degree of the polynomial used to express the forward kinematics. The complexity of solving an SDP is also quadratic in the size of the semidefinite variables and therefore quartic in the the polynomial degree \cite{jiang2020improved}. 

The size of the semidefinite variables in \eqref{E: cert with ellipse} and \eqref{E: cert with polytope} scale as the square of the degree of the polynomial used to express the forward kinematics. Supposing there are $n$ links in the kinematics chain between $\calA$ and $\calB$, then choosing the $j$th link along the kinematics chain as the reference frame $F$ leads to scaling of order $j^{2} + (n-j)^{2}$. Choosing the reference frame in the middle of the chain minimizes this complexity to scaling of order $\frac{n^2}{2}$ and we therefore adopt this convention in our experiments.

\subsection{Basis Selection} \label{S: Basis Selection}
The condition that a polynomial can be written as a sum of squares can be equivalently formulated as an equality constraint between the coefficients of the polynomial and an associated semidefinite variable known as the Gram matrix \cite{parrilo2004sum}. In general, a polynomial in $k$ variables of total degree $2d$ has ${k+2d} \choose {2d}$ coefficients and requires a Gram matrix of size ${k +d} \choose {d}$ to represent which can quickly become prohibitively large. Fortunately, the polynomials in our programs contain substantially more structure which will allow us to drastically reduce the size of the Gram matrices.

We begin by noting from \eqref{E: gen rat poly} that while both the numerator and denominator of the forward kinematics are of total degree $2n$, with $n$ the number of links of the kinematics chain between frame $A$ and $F$, both polynomials are of \emph{coordinate} degree of at most two (i.e. the highest degree of $s_{i}$ in any term is $s_{i}^2$). We will refer to this basis as $\mu(s)$ which is a vector containing terms of the form $\prod_{i = 1}^{n} s_{i}^{d_{i}}$ with $d_{i} \in \{0,1,2\}$ for all $n^{3}$ possible permutations of the exponents $d_{i}$.

Next, we recall the form of  $\alpha^{F, \calA_{j}}(a_{\calA, \calB}, b_{\calA, \calB}, s)$ from \eqref{E: alpha poly} and \eqref{E: beta poly}. If $a_{\calA, \calB}(s)= a^T_{\calA, \calB}\eta(s)$ and $b_{\calA, \calB}(s) = b^T_{\calA, \calB}\eta(s)$ for some basis $\eta$ in the variable $s$, then $\alpha^{F, \calA_{j}}$ and $\beta^{F, \calA_{j}}$ can be expressed as linear functions of the basis $\gamma(s) = \textbf{vect}(\eta(s) \mu(s)^{T})$ where we use $\textbf{vect}$ to indicate the flattening of the matrix outer product. Concretely, if we choose to make $a_{\calA, \calB}(s)$ and $b_{\calA, \calB}(s)$ linear functions of the indeterminates $s$, then $\eta(s) = l(s) = \begin{bmatrix} 1 & s_{1} & \dots & s_{n}\end{bmatrix}$. Therefore $\alpha^{F, \calA_{j}}$ and $\beta^{F, \calA_{j}}$ can be expressed as linear functions of the basis
\begin{align} 
\gamma(s) = \begin{bmatrix} \mu(s) & s_{1}\mu(s) & \dots & s_{n}\mu(s) \end{bmatrix} \label{E: chosen basis}
\end{align}

After choosing the basis $\eta(s)$, which determines the basis $\gamma(s)$, the equality constraints \eqref{E: Putinar pos side} and \eqref{E: Putinar neg side} constrain the necessary basis needed to express the multiplier polynomials $\lambda(s)$ and $\nu(s)$. The minimal such basis is related to an object known in computational algebra as the Newton polytope of a polynomial $\textbf{New}(f)$ \cite{sturmfels1994newton}. The exact condition is that the  $\textbf{New}(\eta(s)) + \textbf{New}(\mu(s)) \subseteq \textbf{New}(\rho(s)) + \textbf{New}(l(s))$ where the sum in this case is the Minkowski sum.

If we choose $\eta(s)$ as the linear basis $l(s)$, then we obtain the condition that $\textbf{New}(\rho(s)) = \textbf{New}(\mu(s))$ and since $\mu(s)$ is a dense, even degree basis then we must take $\rho(s) = \mu(s)$. Choosing $\eta(s)$ as the constant basis would in fact result in the same condition, and therefore searching for separating planes which are linear functions of the tangent-configuration space does not increase the size of the semidefinite variables. As the complexity of \eqref{E: cert with ellipse} and \eqref{E: cert with polytope} is dominated by the size of these semidefinite variables, separating planes which are linear functions changes does not substantially affect the solve time but can dramatically increase the size of the regions which we can certify.

\begin{remark}
In the case of certifying that the end-effector of a 7-DOF robot will not collide with the base using linearly parametrized hyperplanes, choosing to express conditions \eqref{E: Putinar pos side} and \eqref{E: Putinar neg side} in the world frame with na\"ively chosen bases would result in semidefinite variables of size ${7+7 \choose 7} = 3432$. Choosing to express the conditions according to the discussion in Section \ref{S: Frame Selection} and choosing the basis $\gamma(s)$ from \eqref{E: chosen basis} results in semidefinite matrices of rows at most $2^4 = 16$. 
\end{remark}

\subsection{Parallelization} \label{S: Parallelization}
While it is attractive from a theoretical standpoint to write \eqref{E: cert with ellipse} as a single, large program it is worth noting that can in fact be viewed as $K + 1$ individual SOS and SDP programs, where $K$ is the number of collision pairs in the environment. Indeed, certifying whether pairs $(\calA_{1}, \calA_{2})$ are collision-free for all $s$ in the polytope $\calP$ can be done completely independently of the certification of another pair $(\calA_{1},  \calA_{3})$ as neither the constraints nor the cost couple the conditions of imposed on any pairs. Similarly, the search for the largest inscribed ellipsoid can be done independently of the search for the separating hyperplanes.

Solving the certification problem embedded in \eqref{E: cert with ellipse} as $K$ individual SOS programs has several advantages. First, as written \eqref{E: cert with ellipse} has $2(m+1)K\sum_{i} \abs{\calA_{i}}$ semidefinite variables of various sizes. In the example from Section \ref{S: Pinball} this corresponds to 35,072 semidefinite variables. This can be prohibitively large to store in memory as a single program. Additionally, solving the problems independently enables us to determine which collision bodies cannot be certified as collision-free and allows us to terminate our search as soon as a single pair cannot be certified. Finally, decomposing the problems into subproblems enables us to increase computation speed by leveraging parallel processing.

We note that \eqref{E: cert with polytope} cannot be similarly decomposed as on this step the variables $c_{i}^T$ and $d$ affect all of the constraints. However, this program is substantially smaller as we have fixed $2mK\sum_{i} \abs{\calA_{i}}$ semidefinite variables as constants and replaced them with $2m$ linear variables representing the polytope. This program is much more amenable to being solved as a single program.

\subsection{Seeding the Algorithm} \label{S: Seeding}
It is worth noting that the alternations in Algorithm \ref{A: Bilinear Alternation} must be initialized with a polytope $\calP_{0}$ for which \eqref{E: cert with ellipse} is feasible. In principle, the alternation proposed in Section \ref{S: Bilinear Alternation} can be seeded with an arbitrarily small polytope around a collision-free seed point. This seed polytope is then allowed to grow using Algorithm \ref{A: Bilinear Alternation}. However, this may require running several dozens of iterations of Algorithm \ref{A: Bilinear Alternation} for each seed point which can become prohibitive as the size of the problem grows. It is therefore advantageous to seed with as large a region as can be initially certified.

Here we discuss an extension of the {\sc{Iris}} algorithm in \cite{deits2015computing} which uses nonlinear optimization to rapidly generate large regions in C-space. These regions are not guaranteed to be collision-free and therefore they must still be passed to Algorithm \ref{A: Bilinear Alternation} to be certified, but do provide good initial guesses. In this section, we will assume that the reader is familiar with {\sc{Iris}} and will only discuss the modification required to use it to grow C-space regions. Detailed pseudocode is available in Appendix \ref{S: iris pseudocode}

{\sc{Iris}} grows regions in a given space by alternating between two subproblems: {\sc{SeparatingHyperplanes}} and {\sc{InscribedEllipsoid}}. The {\sc{InscribedEllipsoid}} is exactly the program described in \cite[Section 8.4.2]{boyd2004convex} and we do not need to modify it. The subproblem {\sc{SeparatingHyperplanes}} finds a set of hyperplanes which separate the ellipse generated by {\sc{InscribedEllipsoid}} from all of the obstacles. This subproblem is solved by calling two subroutines {\sc{ClosestPointOnObstacle}} and {\sc{TangentPlane}}. The former finds the closest point on a given obstacle to the ellipse, while the latter places a plane at the point found in {\sc{ClosestPointOnObstacle}} that is tangent to the ellipsoid.

The original work in \cite{deits2015computing} assumes convex obstacles which enables {\sc{ClosestPointOnObstacle}} to be solved as a convex program and for the output of {\sc{TangentPlane}} to be globally separating plane between the obstacle and the ellipsoid of the previous step. Due to the non-convexity of the C-space obstacles in our problem formulation, finding the closest point on an obstacle exactly becomes a computationally difficult problem to solve exactly \cite{ferrier2000computation}. Additionally, placing a tangent plane at the nearest point will be only a locally separating plane, not a globally separating one.

To address the former difficulty, we formulate {\sc{ClosestPointOnObstacle}} as a nonlinear program. Let the current ellipse be given as $\calE = \{Qu + s_{0}\mid \norm{u}_2 \leq 1 \}$ and suppose we have the constraint that $s \in \calP = \{s \mid Cs \leq d\}$. Let $\calA$ and $\calB$ be two collision pairs and ${}^{\calA}p_{\calA}, {}^{\calB}p_{\calB}$ be some point in bodies $\calA$ and $\calB$ expressed in some frame attached to $\calA$ and $\calB$. Also, let ${}^{W}X^{\calA}(s)$ and ${}^{W}X^{\calB}(s)$ denote the rigid transforms from the reference frames $\calA$ and $\calB$ to the world frame respectively. We remind the reader that this notation is drawn from \cite{tedrakeManip}. The closest point on the obstacle subject to being contained in $\calP$ can be found by solving the program
\begin{subequations}
\begin{gather} 
\min_{s, {}^{\calA}p_{\calA}, {}^{\calB}p_{\calB}} (s - s_{0})^TQ^TQ(s-s_{0}) \subjectto \\
{}^{W}X^{\calA}(s){}^{\calA}p_{\calA} = {}^{W}X^{\calB}(s){}^{\calB}p_{\calB} \label{E: same point constraint}\\
Cs \leq d
\end{gather}\label{E: closest point}
\end{subequations}
This program searches for the nearest configuration in the metric of the ellipse such that two points in the collision pair come into contact. We find a locally optimal solution $(s^{\star},  {}^{\calA}p_{\calA}^{\star}, {}^{\calB}p_{\calB}^{\star})$ to the program using a fast, general-purpose nonlinear solver such as {\sc{SNOPT}} \cite{gill2005snopt}. The tangent plane to the ellipse $\calE$ at the point $s^{\star}$ is computed by calling {\sc{TangentPlane}} then appended to the inequalities of $\calP$ to form $\calP'$. This routine is looped until \eqref{E: closest point} is infeasible at which point {\sc{InscribedEllipse}} is called again.

Once a region $\calP  =\{s \mid Cs \leq d\}$ is found by Algorithm \ref{A: SNOPT IRIS}, it will typically contain some minor violations of the non-collision constraint. To find an initial, feasible polytope $\calP_{0}$ to use in Algorithm \ref{A: Bilinear Alternation}, we search for a minimal uniform contraction $\delta$ of $\calP$ such that $\calP_{\delta} = \{s \mid Cs \leq d - \delta*1\}$ is collision-free. This can be found by bisecting over the variable $\delta \in [0, \delta_{\max}]$ and solving repeated instances of \eqref{E: cert with ellipse}.

Seeding Algorithm \ref{A: Bilinear Alternation} with a $\calP_{0}$ as above can dramatically reduce the number of alternations required to obtain a fairly large region and is frequently faster than seeding Algorithm \ref{A: Bilinear Alternation} with an arbitrarily small polytope.

\section{Supplementary Algorithms} \label{S: iris pseudocode}

We present a pseudocode for the algorithm presented in Appendix \ref{S: Seeding}. A mature implementation of this algorithm can be found in \href{https://github.com/RobotLocomotion/drake/blob/2f75971b66ca59dc2c1dee4acd78952474936a79/geometry/optimization/iris.cc#L440}{Drake}\footnote{\url{https://github.com/RobotLocomotion/drake/blob/2f75971b66ca59dc2c1dee4acd78952474936a79/geometry/optimization/iris.cc\#L440}}.

\begin{algorithm}
\caption{
Given an initial tangent-configuration space point $s_{0}$ and a list of obstacles $\calO$, return a polytopic region $\calP = \{s \mid Cs \leq d\}$ and inscribed ellipsoid $\calE_{\calP} = \{s \mid Qu + s_{c}\}$ which contains a substantial portion of the free C-space (but is not guaranteed to contain no collisions)
}\label{A: SNOPT IRIS}
\SetAlgoLined
 \LinesNumbered
  \SetKwRepeat{Do}{do}{while}
 $(C, d) \gets $ plant joint limits
 \\
 $\calP_{i} \gets \{s \mid Cs \leq d\}$
 \\
  $\calE_{\calP_{0}} \gets $ {\sc{InscribedEllipsoid}}$(\calP_{0})$
 \\
 $j \gets$ number of rows of $C$
 \\
 \Do{$\left(\textbf{vol}(\calE_{i}) - \textbf{vol}(\calE_{i-1}) \right)/ \textbf{vol}(\calE_{i-1})  \geq$ tolerance}{
 \Do{{\sc{FindClosestCollision$(\calP_{i}, \calE_{\calP_{i}})$}} is feasible}
 {
 $(s^{\star},  {}^{\calA}p_{\calA}^{\star}, {}^{\calB}p_{\calB}^{\star}) \gets$ {\sc{FindClosestCollision}}$(\calP_{i}, \calE_{\calP_{i}})$
 \\
 $(c_{j+1}^T, d_{j+1}) \gets$ {\sc{TangentHyperplane}}$(s^{\star}, \calE_{\calP_{i}})$
 \\
 $C \gets \textbf{hstack}(C, c_{j+1}^T)$
 \\
 $d \gets \textbf{hstack}(d, d_{j+1})$
 \\
 $\calP_{i} \gets \{s \mid Cs \leq d\}$
 \\
 $j \gets j +1$
 }
  $\calE_{\calP_{i}} \gets $ {\sc{InscribedEllipsoid}}$(\calP_{i})$
 $i \gets i+1$
 }
 \Return $(\calP_{i}, \calE_{\calP_{i}})$
\end{algorithm}

\putbib
\end{bibunit}

\end{document}